\documentclass[journal]{IEEEtran}

\usepackage{times}
\usepackage{epsfig}
\usepackage{graphicx}
\usepackage{amsmath}
\usepackage{amssymb}
\usepackage{array}
\usepackage{multirow}
\usepackage{epsfig}
\usepackage{booktabs}
\usepackage{tabularx}
\usepackage{color}
\usepackage{subfigure}

\newcommand{\rev}[1]{{#1}}
\newcommand{\rrev}[1]{{#1}}

\begin{document}
\title{StructVIO : Visual-inertial Odometry with Structural Regularity of Man-made Environments}
%
%
%

\author{
Danping Zou$^{1*}$, Yuanxin Wu$^{1}$, Ling Pei$^{1}$, Haibin Ling$^{2}$ and Wenxian Yu$^{1}$

\thanks{$^{1}$Shanghai Key Laboratory of Navigation and Location-based Services, Shanghai Jiao Tong University,  China}
\thanks{ $^{2}$HiScene Information Technologies, China / Temple University, USA}
\thanks{{\tt \small *E-mail: dpzou@sjtu.edu.cn}}
}

\maketitle

\begin{abstract}
We propose a novel visual-inertial odometry approach that adopts structural regularity in man-made environments. Instead of using Manhattan world assumption, we use Atlanta world model to describe such regularity.  An Atlanta world is a world that contains multiple local Manhattan worlds with different heading directions. Each local Manhattan world is detected on-the-fly, and their headings are gradually refined by the state estimator when new observations are coming. With fully exploration of structural lines that aligned with each local Manhattan worlds, our visual-inertial odometry method becomes more accurate and robust,  as well as more flexible to different kinds of complex man-made environments.
Through extensive benchmark tests and real-world tests, the results show that the proposed approach  \rev{outperforms existing visual-inertial systems in large-scale man-made environments}.
\end{abstract}


%
\IEEEpeerreviewmaketitle

\section{Introduction}
\rev{Accurately estimating} the sensor position and orientation in indoor scenes is a challenging problem, which is however a basic requirement for many applications such as autonomous parking, AGVs, UAVs, and Augmented/Virtual reality. Usually SLAM (Simultaneous Localization and Mapping) techniques are applied to solve such problem. Among all SLAM techniques, visual SLAM are the most favorable one to be employed in those systems where the cost, energy, or weight are limited.

  A large number of methods have been developed in the past decade to solve the SLAM problem using video cameras. Some of them exhibit impressive results in both small scale \cite{klein2007parallel} and large scale scenes \cite{mur2015orb}, even in dynamic environments \cite{zou2013coslam}.  With extra measurement data from inertial measurements units (IMUs), so-called visual-inertial odometry (VIO) systems \cite{mourikis2007multi}\cite{leutenegger2015keyframe}\cite{qin2017vins} achieve remarkably better accuracy and robustness than pure vision systems.

Most existing approaches focus on handling general scenes, with less attention to particular scenes, like man-made environments. Those environments exhibit strong structural regularities, where most of them can be abstracted as a box world, which is known as Manhattan worlds
\cite{coughlan1999manhattan}. In such kind of worlds, planes or lines in perpendicular directions are predominant. Such characteristic have been applied to 
indoor modeling \cite{furukawa2009manhattan}, scene understanding \cite{gupta2010blocks} and heading estimation \cite{ruotsalainen2012mitigation}. With the help of Manhattan world assumption, the robustness and accuracy of visual SLAM can been improved as shown in \cite{zhou2015structslam}.

However, the Manhattan world assumption is restrictive to be applied to general man-made scenes, which may include oblique or curvy structures. Most common scenes are those that contain multiple box worlds, each of which have different headings. If the detected line features are forced to be aligned with a single Manhattan world in such cases, the performance may become worse. In this work, we try to address this issue and extend the idea of exploration of structural regularity of man-made scenes \cite{zhou2015structslam} to visual-inertial odometry (VIO) systems. 

The key idea is to model the scene as an Atlanta world \cite{schindler2004atlanta} rather than a single Manhattan world. An Atlanta world is a world that contains multiple local Manhattan worlds, or a set of box worlds with different orientations. Each local Manhattan world is detected on-the-fly, and their headings are gradually refined by the state estimator when new observations are coming. The detected local Manhattan world is not necessary a real box world. As we will see, we allow a local Manhattan world being detected even if the lines are found to be aligned with only a single horizontal direction. It enables our algorithm flexible to irregular scenes with triangular or polygonal shapes.

The benefit of using such structural regularity in VIO system is apparent :  \rev{The horizontal lines aligned with one of the local Manhattan worlds give rise a constraint in heading, making the orientation error without growing in this local area; Even though no Manhattan world has been detected, 
a vertical line indicates the gravity direction and immediately renders the roll and pitch of the sensor pose observable.} 

Based on the above mentioned ideas, we present a novel visual-inertial odometry method, which is built upon the state-of-the-art VIO framework and made several extensions to incorporate the structural information, including structural line features and local Manhattan worlds. We describe the extensions in detail, including the Atlanta world representation, structural line parameterization, filter design, line initialization, tracking and triangulation of line tracks, and detection of Manhattan worlds. 


We have conducted extensive experiments on both public benchmark datasets and challenging datasets collected at different buildings. The results show that our method, 
incorporating the structural regularities of the man-made environments through exploring both structural features and multiple Manhattan world assumption, achieves \rev{the best performance in all tests.} 
We highlight major technical contributions as follows.

1)	A novel line representation with minimum number of parameters seamlessly integrates the Atlanta world assumption and structural features of the man-made buildings. Each local Manhattan world is represented by a heading direction and refined by the estimator as a state variable.

2)	Structural  lines (lines with dominant directions) and the local Manhattan world are automatically recognized on-the-fly. If no structural line has been detected, our approach works just like a point-based system. Note that when no Manhattan world has been detected, vertical lines still help in estimation if they can be found. This makes our method flexible to different kinds of scenes besides indoor scenes.

3)	We also made several improvements on the estimator and line tracking. A novel information accumulation method handles the dropped measurements of long feature tracks and enables better feature estimates; a line tracking method by sampling multiple points and delayed EKF update makes the tracker more reliable.


\section{Related work}

The structural regularity of man-made environments is well known as Manhattan world model \cite{coughlan1999manhattan}. The first indication from this regularity is that line features are predominant in man-made environments. Researchers have attempted to use straight lines as the landmarks since the early days of visualm SLAM research \cite{smith2006real}\cite{sola2009undelayed}\cite{perdices2014lineslam}. Recent works that try to use straight line features in visual SLAM \cite{pumarola2017pl} and visual-inertial odometry \cite{he2018pl} can also be found. However, most visual SLAM  \cite{klein2007parallel}\cite{mur2015orb}\cite{zou2013coslam} or visual-inertial \cite{mourikis2007multi} \cite{leutenegger2015keyframe}\cite{qin2017vins}  systems  prefer to use only point features. There are several reasons. First, points are ubiquitous features that can be found nearly in any scenes. Second, compared with line features, point features are well studied to be detected easily and tracked reliably. Another reason is that a straight line has more degree of freedom \rrev{(4 DoF)\cite{zhang2014structure}} than a single point (3 DoF), which makes a line more difficult to be initialized (especially the orientation) and be estimated (usually 6 parameters are required like Plucker coordinates
\cite{sola2012impact}) than a point. It has been shown that adopting line features in a SLAM system may sometimes lead to a worse performance than that of using only points \cite{zhou2015structslam}. Therefore the above mentioned issues need to be carefully addressed, e.g. using stereo camera settings \cite{gomez2016robust} and delayed initialization by waiting multiple key frames \cite{pumarola2017pl}. Nevertheless lines are still a good complement to points, which allow adding extra measurements in the estimator to get more accurate results. This is particularly helpful when there are not enough point features in some texture-less indoor scenes. 

Another indication from structural regularity is that structural lines are aligned with three axes of the coordinate frame of a Manhattan world. The directional information encoded in the lines offers a clue about the camera orientation, which appears as vanishing points in the image. The vanishing points from parallel lines on the images relates to the camera orientation directly. It has shown that using vanishing points can improve visual SLAM \cite{lee2009vpass}\cite{zhang2012loop} and visual-inertial odometry \cite{camposeco2015using}. However, in those methods the line features are used as only intermediate results for extracting vanishing points. After that,  lines are simply discarded. It should be helpful by integrating them in the estimator in the same way as points.

Most existing methods explore only the partial information of structural regularity - they either use straight lines without considering their prior orientation, or use the prior orientation without putting lines as extra measurements for better estimation.
A few of existing methods consider both aspects \cite{zhou2015structslam}\cite{kottas2013exploiting}. In \cite{zhou2015structslam}, the lines with  prior orientation are named as structural  lines and treated them as landmarks the same as point features for visual SLAM. The method \cite{kottas2013exploiting} has a similar spirit but puts focus on visual-inertial odometry. The assumption of only three dominant directions of those methods limit their application to simple scenes that contain no oblique structure.
Both methods rely on rigid initializations to detect three directions, requiring at least one vanishing point (in horizontal direction) to be captured in the image for visual-inertial systems and two vanishing points for pure vision systems before the algorithm can start.

In this work, we take a step further to present a powerful visual-inertial odometry method by addressing above mentioned issues. We propose to use Atlanta world  \cite{schindler2004atlanta} to allow multiple Manhattan worlds with different directions, and detect each Manhattan world on-the-fly and refine their headings in the state gradually. The proposed method does not need to capture any vanishing points at the initialization stage. Our novel line parameterization anchors each line to a local Manhattan world, which reduces the degree of freedom of the line parameter and enables line directions being refined along with the heading of the Manhattan world as more evidences are collected.

\section{structural  lines and Atlanta worlds}
\label{sec:structure_line}
\begin{figure}
 \centering
\includegraphics[width=0.90\linewidth]{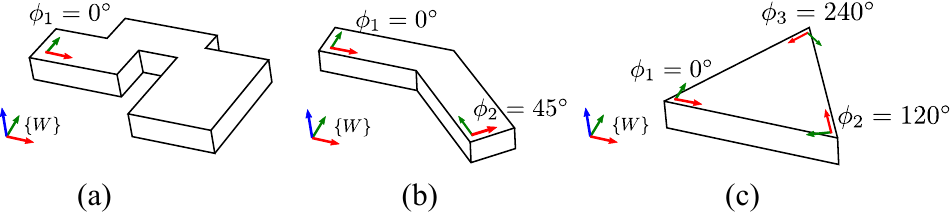}
\caption{(a) A Manhattan world ; (b) An Atlanta world consists of two local Manhattan worlds with the heading directions $0^\circ$ and $45^\circ$ respectively (c) An Atlanta world consists of three local worlds indicated by three directions $0^\circ,120^\circ,$ and $240^\circ$. Those worlds are however not real box worlds.}
 \label{fig:structure_lines}
 \end{figure}

To better model  general man-made scenes, we adopt the Atlanta world \cite{schindler2004atlanta} assumption in our approach. It is  an extension of the Manhattan world  assumption - the world is considered as a superposition of multiple Manhattan worlds with different horizontal orientations $\phi_i \in [0,\pi/2), (i=1,2,\ldots,N)$ as shown in Figure \ref{fig:structure_lines}. Note that 
each local  world is not necessary a real box world containing three perpendicular axes. One horizontal direction can determine one local world as shown in Figure \ref{fig:structure_lines}(c). \rev{This allows to model irregular scenes with triangular or polygonal shapes. We also set a dummy Manhattan world $\phi_0$, whose orientation is the same as that of the global world $\{W\}$.}

We establish the global world coordinate system $\{W\}$ with $Z$-axis pointing up (reverse direction of gravity)  on the location where odometry starts. The IMU coordinate system and the camera coordinate system are denoted by $\{ I\}$ and $\{C\}$. Their poses are described  by $^W_I\tau = (^W_C{q}, {^W{p}_I})$ and $^W_C\tau = (^W_C{q}, {^W{p}_C})$. Here ${^W_I{q}},{^W_C{q}}$ are rotation transformations represented in unit quaternions and their matrix forms are ${^W_IR}$ and $^W_CR$ respectively. ${^Wp_I} $ and ${^Wp_C}$ are the origin of the IMU frame and the origin of the camera frame expressed in the world coordinate system. 


Each line is anchored to the local coordinate system where the line is firstly detected on the image. We call the anchored coordinate system  \emph{starting frame}, denoted by $\{S\}$, \rrev{where the orientation
is the same as that of the local
Manhattan world $\phi_i$ related to this line, and the origin is the camera position when the line being firstly detected.} \rev{The origin of the starting frame, $^Wp_S$, will be changed to a new position in the state and updated by the filter as we will see in Sec. \ref{sec:marg}.}

\begin{figure}
\centering
\includegraphics[width=0.9\linewidth]{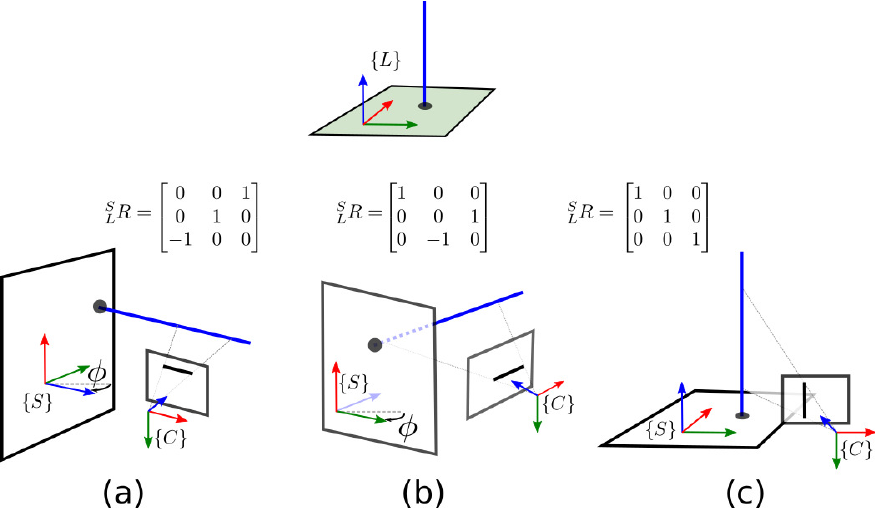}
\caption{Starting frames of different structural lines (aligned with $X, Y,Z$ axes) transformed from the parameter space $\{L\}$. The camera captures those lines in a novel view denoted by the camera frame $\{C\}$. Note that the staring frame is aligned with some local Manhattan world whose heading is rotated about $\phi$ from the world frame. }
\label{fig:structural_lines}
\end{figure}

 For a given line attached to the starting frame $\{S\}$, we can find a rotation $^S_LR$ , that transforms this line from a parameter space $\{L\}$ into $\{S\}$, where the line is aligned with the Z axis of $\{L\}$ as shown in Figure \ref{fig:structural_lines}. In the parameter space $\{L\}$, the line can be simply represented by the intersection point on  the XY plane, namely \rrev{$ ^Ll_p = (a,b, 0)^T$.}  Here we use the inverse depth representation of the intersection point to describe a line, namely, $ (\theta,\rho,0)^T$, where $\rho = 1/\sqrt{a^2+b^2}$ and $\theta = \text{atan2}(b,a)$. The inverse depth representation is known as a good parameterization that can describe infinite features and minimize nonlinearity in feature initialization \cite{civera2008inverse}\cite{zhou2015structslam}.

 
  The line in the starting frame is computed from a rotation transformation $^S_LR$ and the intersection point $^Ll_p$, 
  \begin{equation}
\begin{split}
  {^S_LR}\,l_p = &a\, ^S_LR(:,1)  + b ^S_LR(:,2) \\
  = &\frac{\cos\theta}{\rho}\,^S_LR(:,1)+\frac{\sin\theta}{\rho}\,{^S_LR(:,2)}
\end{split}
  \label{eq:sl}
  \end{equation}
  
  For structural lines that are aligned with any axis of the three axes of the local Manhattan world, the rotation $^S_LR$ is one of the following constants:
 \begin{equation}\small  \left[ \begin{array}
{ccc} 0& 0 & 1 \\ 0 & 1 & 0 \\ -1 & 0 & 0\end{array}\right]
,  \left[ \begin{array} {ccc} 1& 0 &0 \\ 0 & 0 & 1 \\ 0 &-1 & 0\end{array}\right] \text{and}
\left[\begin{array} {ccc} 1& 0 &0 \\ 0 & 1 & 0 \\ 0 & 0 & 1\end{array} \right],
\end{equation}
which correspond to lines aligned with $X,Y,Z$ axes.  

The transformation from the starting frame $\{S\}$ to the world frame $\{W\}$ is determined by the rotation $^W_SR(\phi_i)$ and the camera position $^Wp_s$, where $^W_SR(\phi_i)$ is a rotation about the gravity direction, namely ,
\begin{equation}
\small
 ^W_SR(\phi_i) = \left[ \begin{array}{ccc} \cos(\phi_i) & \sin(\phi_i) & 0 \\
-\sin(\phi_i) & \cos(\phi_i) & 0 \\ 0 & 0 & 1 \end{array} \right].
\label{eq:phi}
\end{equation}


\rev{For vertical lines, the axes of their starting frame are the same as those of the world frame, $^W_SR = I_{3 \times 3}$. We use $^W_SR(\phi_0), (\phi_0 = 0)$ to represent this kind of starting frames.}

To obtain the projection of a structural line on the image, it requires to project both the
intersection point \rrev{$^Ll_p$} and  the $Z$ direction in the parameter space onto the image plane.
The coordinates of the intersection point \rrev{$^Ll_p$} in the world frame is computed as
\begin{equation}
^Wl_p =  {^W_SR(\phi_i)} {^S_LR}\, {^Ll_p} + {^Wp_S},
\label{eq:wl}
\end{equation}
which can be further transformed into \rev{ a camera frame, $\{C\}$, by}
\begin{equation}
{^Cl_p} = {^C_WR}\, {^Wl_p} + {^Cp_W},
 \label{eq:cw}
 \end{equation}
where $(^C_WR,{^Cp_W})$ represents the transformation from the world frame to the camera frame. From  (\ref{eq:sl}) (\ref{eq:phi}) (\ref{eq:wl}) (\ref{eq:cw}), by replacing \rrev{$^Ll_p$} with the inverse depth representation $ (\theta, \rho,0)^T$, we get the homogeneous coordinates of the 2D projection of  the intersection point on the image plane
\begin{equation}
\rrev{{^Cl_p }} \sim {^C_WR}\,{{^W_SR}(\phi_i)} {^S_LR}\cdot r + ({^C_WR}\, {^Wp_S}+{^Cp_W}) \cdot \rho,
\end{equation}
where $ r = [\cos\theta, \sin\theta,0]^T$.
The \rev{homogeneous coordinates} of the vanishing point projected by the $Z$ direction of the parameter space  \rev{are} computed as:
 \begin{equation}
 \rev{^Cv} \sim {^C_WR } \,\, {^W_SR(\phi_i)} {^S_LR}(:,3).
 \label{eq:vanishing_point}
 \end{equation}
Here $^S_LR(:,3)$ is the third column of $^S_LR$. 
Taking the camera intrinsics ($K \in \mathbb{R}^{3\times 3} $) into account,  we get the line equation on the image by:
 \begin{equation}
\rrev{^{im}l} = (K^{-T}) ( \rev{{^Cl_p \times {^Cv} }}).
\label{eq:line_image}
 \end{equation}
From above definitions, we are able to establish the relationship between the 3D line and its 2D observations  given the two parameters of the  inverse depth representation $l=(\theta,\rho)^T$,  the Manhattan world $\phi_i$ which the line lies in, and
the direction ($X$,$Y$,$Z$) to which the line belongs \rev{(described by $^S_LR$)}. 

The line projection can be written as a function
\begin{equation}
\rrev{^{im}l} = \Pi(l,\phi_i,{^S_LR},{^W_C\tau}),
\end{equation} where ${^W_C\tau} = (^W_Cq, ^Wp_c)$ denotes the camera pose.
If we use the IMU pose, $^W_I\tau$, instead of the camera pose, we have
\begin{equation}
\rrev{^{im}l} = \Pi(l,\phi_i,{^S_LR},{^I_C\tau},{^W_I\tau}),
\end{equation}
where ${^I_C\tau}$ represents the relative pose between the IMU and camera frames, \rev{and can be included in the filter for update to account for inaccurate calibration.}

 \begin{figure}[h]
 \centering
\includegraphics[width=0.9\linewidth]{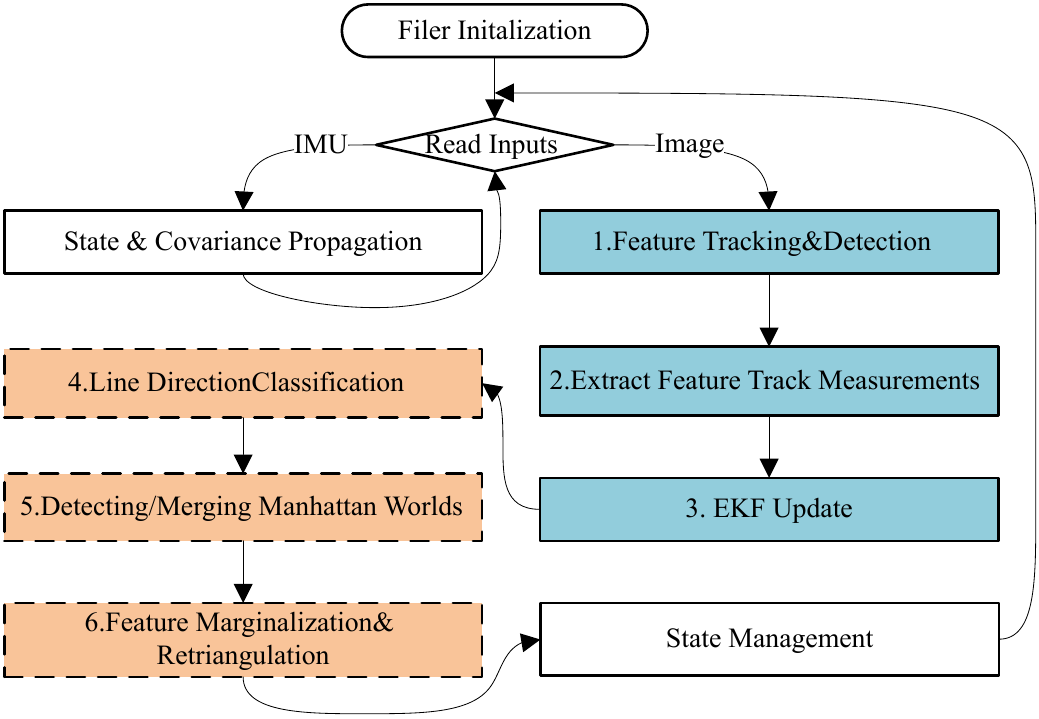}
\caption{Flowchart of StructVIO. Solid boxes represents the key components of a typical point-only approach and among all of them, (1-3) are required to be extended to adopt structural line features. Dash boxes (4-6) are  novel components involved in StructVIO. }
 \label{fig:pipeline}
 \end{figure}
\section{System overview}
As shown in Figure \ref{fig:pipeline}, we adopt the EKF framework in our visual-inertial system.
The state vector of our filter is defined as the following \footnote{We switch the state into a row or column vector accordingly for brevity in the following text.}:
\begin{equation}
\small
x_k = [{{x}_{I_k}}, {^I_C\tau}, \phi_1,\cdots \phi_N, {^W_{I_1}\tau},\cdots  {^W_{I_M}\tau}],
\end{equation}
where ${{x}_{I_k}}$ indicates the IMU state at time step $k$, including its pose, velocity,  and the  biases of the gyroscope and the accelerometer, 
\begin{equation}
{{x}_{I_k}} = [^W_{I_k}\tau, {^Wv_{I_k}}, b_{g_k}, b_{a_k}],
\end{equation}
where $^W_{I_k}\tau = [^W_{I_k}q, {^Wp_{I_k}}]$ is the IMU pose at the $k$-th time step.
 We also put the relative pose between the IMU and camera frame $^I_C\tau = [^I_{C}q, {^Ip_C}]$ into the filter to allow it to be updated if sometimes inaccurate calibration is presented.

By adopting Atlanta world model, we detect each box world (or local Manhattan world) on the fly and include the heading of each local world in the state $\phi_i,(i=1, \ldots N)$. Those headings will be refined gradually when more observations becomes available.

The historical IMU poses are ${^W_{I_i}\tau} = [^W_{I_i}q,{^Wp_{I_i}}], (i = 1,\ldots M)$. Those historical IMU poses are cloned from the current IMU state $^W_{I_k}\tau$ at different time steps. Some of them will be removed from the state if the number of historical poses exceeds the maximum number allowed.

The covariance matrix at the $k$-th time step is denoted by $P^x_k$. Our VIO estimator is to recursively estimate the state $x_k$ and the covariance $P^x_k$ starting from the initial state and covariance, $x_0$ and $P^x_0$.

We follow the approach \cite{mourikis2007multi} to design our filter.
All the features, both points and structural lines, are not included in state. They are estimated separately outside of the filter and only used for deriving geometric constraints among IMU poses in the state. 
The pipeline of our filter is shown in Figure \ref{fig:pipeline}.
we'll present the details of our filter design in following sections. 

\paragraph{Dynamical model}
The dynamical model involves state prediction and covariance propagation. The state is predicted by IMU integration, namely
\begin{equation}
\hat{x}_k \leftarrow [\mathcal{G}(x_{I_k},\omega_{k,k-1}, a_{k,k-1}), {^I_C\tau},\phi_1,\cdots].
\label{eq:state_prediction}
\end{equation}
Here, $\mathcal{G}(\cdot)$ represents IMU integration, where we apply Runge-Kuatta method. To compute the slope values \rev{in Runge-Kutta method} more accurately,  we use measurements  in both $k^{th}$ and $k-1^{th}$ time steps, $\omega_{k,k-1}$ for gyroscope and $a_{k,k-1}$ for accelerometer, to linearly approximate the angular velocity and acceleration inside of the time interval from $k-1$ to $k$.

\rev{The covariance is propagated approximately using the error state\cite{sola2017quaternion}.}
Let $\delta x_k$ be the error state corresponding to the state vector $x_k$.  The predicted error state is  given by
\begin{equation}
\delta \hat{x}_{k} \leftarrow F\delta{x}_{k} + G n_{imu},
\end{equation}
where $F$ is the state transition matrix and $G$ is the noise matrix. The variable $n_{imu}$ represents the random noise of IMU,  including  the measurements noise  and the random walk noise of the biases.  
The covariance of the error state is then computed as
\begin{equation}
\hat{P}^x_{k+1} \leftarrow FP^x_{k}F^T + G Q G^T,
\label{eq:cov_propagation}
\end{equation} where $Q$ is a covariance matrix corresponding to the IMU noise vector $n_{imu}$.
From (\ref{eq:state_prediction}), we know that except the error state of IMU, the error states of other variables remain unchanged. So the transition matrix $F$ has the following form
$
F = \left [\begin{array}{cc} F_{imu} & 0 \\ 0 & I \end{array}\right]
$, where $F_{imu}$ is computed as
\begin{equation}
\small
F_{imu} = \left[\begin{array}{ccccc}
I_{3\times3} & 0 & 0 & -{^W_{I_k}R} \Delta t & 0 \\
0 & I_{3\times3} & I_{3\times3} \Delta t & 0 & 0 \\
0 & 0 & 0 & I_{3\times3} & 0 \\
0 & 0 & 0 & 0 & I_{3\times3}
\end{array}
\right].
\end{equation}
$^W_{I_k}R$ is the rotation matrix corresponding to the quaternion  representing the IMU's orientation,  $^W_{I_k}q$, and $\Delta t$ is the time interval between the $k-1^{th}$ and $k^{th}$ steps.

The noise matrix is computed as 
$
G = \begin{bmatrix}
G_n \\
0
\end{bmatrix}
$, where $G_n$ represents the noise matrix of  IMU measurements
\begin{equation}
\small
G_{n} = \left[
\begin{array}{cccc}
-{^W_{I_k}R}\Delta t & 0 & 0 & 0\\
0 & 0 & 0 & 0 \\
0 & -{^W_{I_k}R}\Delta t & 0 & 0\\
0 & 0 & I_{3\times3} & 0\\
0 & 0 & 0 & I_{3\times3}
\end{array}
\right],
\end{equation}
From (\ref{eq:state_prediction}) and (\ref{eq:cov_propagation}), we are able to predict the current state and propagate the covariance from the last time step.

 \paragraph{Measurement model}
The measurement model involves both point and line features. Here we only describe the measurement model of structural lines, as the measurement model of points is the same as the one described in \cite{mourikis2007multi}. 
To derive the measurement model of structural lines, we need to compute the projection of a given structural line on the image first.

The measurement model of structural lines 
 can be derived from (\ref{eq:line_image}). Let the structural line projected on the image be $\rrev{^{im}l} = (l_1,l_2,l_3)^T$ and the two end points of  the associated line segment  be $s_a, s_b$ (homogeneous coordinates) in the $k$-th view.
We adopt a signed distance to describe the closeness between the  observation (line segment detected in the image) and the predicted line from perspective projection \cite{zhou2015structslam},
\begin{equation}
\small
\begin{split}
r_k = D(s_a,s_b, {^{im}l})
 = D(s_a,s_b,\Pi(l,\phi_\rrev{i}, {^I_C\tau}, {^W_{I_k}\tau})) \\
=\left[\begin{array}{c}
{s}^T_a\, {^{im}l} / \sqrt{l^2_1+l^2_2}\\ 
{s}^T_b\, {^{im}l}/ \sqrt{l^2_1+l^2_2}
\end{array}
\right]
\end{split}
\label{eq:reprojection_error}
\end{equation}
By linearizion about the last estimation of line parameters, the residual $r_k$ in  the $k$-th view is approximated as:
\begin{equation}
\small
r_k = h_0  + J_l \delta l + J_{\phi_i} \delta \rrev{\phi_i} + J_{IC} \delta\, {^I_C\tau} + J_{WI_k} \delta \, {^W_{I_k}\tau}
\end{equation}
where  $J_l \in \mathbb{R}^{2\times2}, \rrev{J_{\phi_i}} \in \mathbb{R}^{2\times1}, J_{IC} \in \mathbb{R}^{2\times 6}, J_{WI_k} \in \mathbb{R}^{2\times6}$ are the Jacobian matrices  with respect to
$l, \phi_{\rrev{i}}, {^I_C\tau}, {^W_{I_k}\tau}$. By stacking the measurements from all visible views, we get the following measurement equation for a single structural line:
\begin{equation}
\small
\begin{split}
z = &H_l \delta l + H_{\rrev{\phi_i}} \delta \rrev{\phi_i} +  H_{CI} \delta \, {^I_C\tau} + H_{WI} [ \delta {^W_{I_1}\tau} \cdots \delta {^W_{I_M}\tau}].
\end{split}
\end{equation}
We then project the residual $z$ to the left null space of $H_l$ to yield a new residual $z^{(0)}$ defined as
\begin{equation}
\small
\begin{split}
z^{(0)} = & H^{(0)}_{\rrev{\phi_i}} \delta \rrev{\phi_i} + H^{(0)}_{CI} \delta \, {^Cx_I}   + H^{(0)}_{WI} [ \delta {^W_{I_1}\tau}\cdots \delta {^W_{I_M}\tau}],
\end{split}
\label{eq:null_space_measurement}
\end{equation}
We write it in a brief form: 
\begin{equation}
z^{(0)} = H^{(0)} \delta x.
\end{equation} 
By doing this, structural lines are decoupled from state estimation, significantly reducing the number of variables in the state.

Note that unlike points, the measurement model of structural lines has a novel part related to the horizontal direction ($\phi_i$) of a given Manhattan world. That means the horizontal direction of a Manhattan world can be estimated by the filter, allowing us to use multiple box worlds to model the complicate environments.

In our implementation, we adopt numerical differentiation to compute all those Jacobian matrices, as analytic forms are too complicated to be computed. By taking the measurement noise into account,  we have $z^{(0)} \sim \mathcal{N}(0, \sigma_{im} I)$, where $\sigma_{im} \in \mathbb{R}^+$ describes the noise level of the line measurement.

\paragraph{EKF update}
 There are two events to trigger EKF updates.  The first one is a structural line being tracked is no longer detected in the image. All the measurements of this structural line are used for EKF update and the historical poses where the line is visible are involved in computation of the measurement equation (\ref{eq:null_space_measurement}). To account for occasional tracker failure, we do not trigger EKF immediately, but wait until the tracker is unable to recover for a number of frames.  This delayed update strategy significantly improves the performance of line tracking as we observed in tests. 

The second event that triggers EKF update is that the number of poses in the state exceeds the maximum number allowed. In such case, we select one-third poses evenly distributed  in the state starting from the second oldest pose, and use all the features including both points and lines visible in those poses to construct the measurement equation similar to the approach described in \cite{mourikis2007multi}.

The EKF update process follows the standard procedure, where Joseph form is used to update the covariance matrix to ensure numerical stability.

\paragraph{State management}
State management involves adding new variables to the state and removing old variables from the state. Adding new variables to the state, or state augmentation, is caused by two events.
The first one is a new image has been received. In this case, the current pose of IMU is appended to the state and the covariance matrix is also augmented, 
\begin{equation}
x_k \leftarrow 
J_I
x_k = \begin{bmatrix}
x_k\\
_{I_k}^W\tau\\

\end{bmatrix}
 \, \text{and} \, P^x_k \leftarrow J_I P^x_k J^T_I,
\end{equation}
where $J_I$ represents the operation of cloning and insertion of IMU variables. The second event is about a new Manhattan world has been detected in the environment.
Let the heading of the newly detected Manhattan word be $\phi$.
Similarly, we have
\begin{equation}
x_k \leftarrow  \begin{bmatrix}
x_k \\
\phi
\end{bmatrix}\,\text{and}\,P^x_k \leftarrow \begin{bmatrix} 
P^x_k & P^{x\phi}_k\\
P^{\phi x}_k & P^\phi_k \\
\end{bmatrix}
\end{equation}
Note that the uncertainty of the heading $\phi$ depends on many factors in the process of Manhattan world detection, which involves pose estimation, line detection, and calibration error. 
Though we can compute Jacobian matrices with respect to all related error factors to get an accurate $P^{x\phi}$ and $P^\phi_k$, we found that it works well by simply neglecting the correlation between
$x$ and $\phi$, $P^{x\phi}_k = 0$, and treating $P^\phi_k$ as a preset constant. In our implementation, we let $P^\phi_k = \sigma^2_\phi$,  where $\sigma_\phi=5^\circ$.

 \section{Implementation of Key components}
\subsection{Line detection \& tracking}
\label{sec:feature_detection}
  We use the LSD line detector \cite{von2010lsd} to detect line features in the image and use
3D-2D matching to track the line measurements for each structural line. The advantage of 3D-2D tracking is that it utilizes the camera motion to predict possible position of the structural line on the image to reduce the range for searching correspondences. Another advantage is that it is more  convenient  for handling occasional  tracker lost.
\begin{figure}
 \centering
\includegraphics[width=1.0\linewidth]{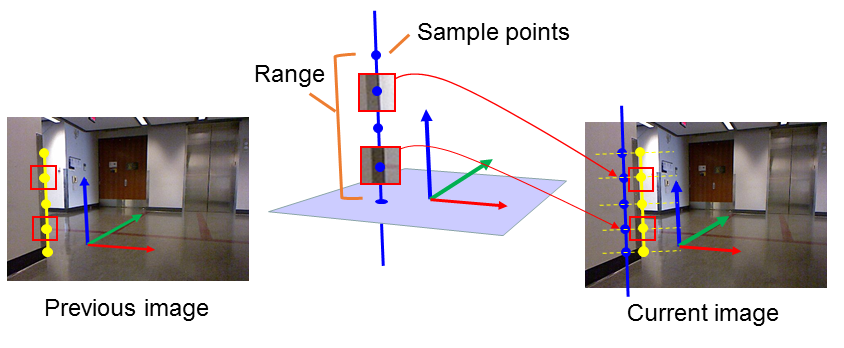}
\caption{Illustration of structural line tracking by matching sample points}
 \label{fig:line_association}
 \end{figure}
For each structural line, apart from those geometric parameters introduced in Sec. \ref{sec:structure_line}, we also introduce a variable $r = (r_s, r_e)^T $ to represent
  the range of  the structural line in 3D space.  The two end points of the structural line correspond to  $\rrev{^Ll_{s}} = (a, b, r_s)^T$ and $\rrev{^Ll_e} = (a,b,r_e)^T$ in the parameter space $\{L\}$.
When a new image arrives,  the structural line is projected onto the image  using the predicted camera pose by IMU integration. The next step involves searching line segments detected in the new image that are close to the line projection.

This can be done by checking the positional and directional difference between the line projection and the detected line segment. 
We then get a  set of candidate line segments. After that, we attempt to figure out the real line correspondence among those candidates based on image appearance. Instead of considering only the image appearance around the middle point of a line segment as described in \cite{zhou2015structslam},  we \rev{consider using} multiple points on the line to improve the tracking robustness.

We sample those points  on the structural line
  by dividing the range into equally distributed values: $r_s, r^{(1)},r^{(2)},\ldots,r^{(K-2)}, r_e$. For each sample point, we keep its image patch around its projection on the last video frame, and search its corresponding points on the candidate line segments by ZNCC  image matching with a preset threshold (ZNCC $> 0.6$), as shown in Figure \ref{fig:line_association}. Finally, we choose the line segment that has the largest number of corresponding
  points as the associated one.  The proposed two-phase line tracking method is proofed to be very effective through extensive tests. 

 \subsection{\rrev{Recognition of structural line segments}}
 \label{sec:line_classification}
\rev{Among the line segments that newly extracted in the image, we attempt to recognize the structural ones, those aligned with three axes of a Manhattan world, and classify them into different directions.} For \rev{each structural line segment}, we also try to figure out in which local Manhattan world they lie. In the first step, we compute all the vanishing points related to \rev{three directions of a Manhattan world.} 
 From (\ref{eq:vanishing_point}), the vanishing point of $Z$ direction is 
\begin{equation}
\rrev{v_z} = K \,\, ^C_WR\,\, \begin{bmatrix}
0 & 0 &1 \\
\end{bmatrix}^T,
\label{eq:vanishing_point_z}
\end{equation} 
where $K$ is the camera intrinsic matrix and $^C_WR\rev{={(^W_CR)^T}}$ represents \rev{the current orientation of the camera}. 
 Similarly, for $X,Y$ directions, we have
\begin{equation}
\rrev{v^{\phi_i}_x} = K\,\, ^C_WR\,\, \begin{bmatrix}
 \cos\phi_i,
 -\sin\phi_i,
 0
  \end{bmatrix}^T
\label{eq:vanishing_point_x}
\end{equation} and
\begin{equation}
\rrev{v^{\phi_i}_y} = K\,\, ^C_WR\,\, \begin{bmatrix}
\sin\phi_i,
\cos\phi_i,
 0
 \end{bmatrix}^T.
\label{eq:vanishing_point_y}
\end{equation}
Note that, only the vanishing points of horizontal directions depend on the heading $\phi_i$ of the local Manhattan world. We can therefore recognize the vertical lines even if there is no local Manhattan world being detected.

To recognize the structural ones \rev{from all detected line segments}, we draw a ray  from each vanishing point to the middle point of line segment $S$, \rev{and} then check the consistency between the ray and the line segment $S$, including the closeness and the directional difference.
We set thresholds for the closeness and the directional difference for evaluating the consistency, and evaluate all vanishing points for each line segment.
The line segment is recognized as a structural one if it is consistent with one of those vanishing points. \rev{The Manhattan world related to this segment is then determined from the corresponding vanishing point.}

\rev{The vanishing point with the best consistency is chosen when sometimes there are multiple consistent vanishing points.}
The remaining line segments that are not consistent with any vanishing points 
are simply excluded in our state estimator.  
Note that if no Manhattan world has been already detected, vertical structural line segments can still be recognized using the vanishing point related to the vertical direction (\rev{with a dummy Manhattan world $\phi_0$ assigned}) as \rev{(\ref{eq:vanishing_point_z}) shows.}
	

\subsection{Initialization of structural lines} 
After recognizing structural segments \rev{among newly extracted line segments} in the current image, we choose \rev{only some of them for initialization} to avoid redundant initialization (multiple line segments from a single 3D line are initialized) and let the initialized lines be well distributed in the image. 

We found the following two rules work well for selecting informative line segments for initialization: 1) the line segments are among the longest ones; 2) the line segments are not close to those segments already initialized. 

Following the above rules, we firstly remove  line segments close to those  being already initialized. \rev{Next, we} sort the remaining line segments by their length in decreasing order and put them in a queue, and then use the following iterations to select line segments for initialization:

 1) pop the line segment $s$  at the head of queue and remove it from the queue;

  2)  initialize a new structural line from $s$;

  3) remove the line segments in the queue that are close to $s$ and goto Step 1 until the queue is empty or the number of structural lines has reached the maximum number allowed.

The remaining issue is to initialize new structural lines from the chosen segments. The key to initialize a new structural line $l = (\theta,\rho)$ is to find the angular parameter $\theta$, while the inverse depth value $\rho$ can be set to a preset value.
The first step of initialization is to establish a starting frame for the structural line. For all structural lines in vertical directions, we choose the starting frame as the one whose axes are aligned with the world frame $\{W\}$, or a virtual Manhattan world that $\phi_0 = 0$. This choice  makes it convenient to represent vertical lines if no Manhattan world has been detected. For structural lines in horizontal direction, the starting frame is selected as the one whose axes are aligned with the local Manhattan world $\phi_i$. 

The angular parameter $\theta$ is determined by the direction from the camera center to the line on the $XY$ plane of the local parameter space $\{L\}$. The direction can be approximated by the ray from the camera center to \rev{the middle point} on the line segment $s$.

Let $m$ (in homogeneous coordinates) be the middle point of $s$. The back projection ray of $m$ in the camera frame is $K^{-1}m$, which is transformed into the local parameter space $\{L\}$ by 
\begin{equation}
{^Lm} = {^L_SR}\,\,{^S_WR(\phi_i)}\,\,{^W_CR}\,\,K^{-1}m.
\end{equation}
Since $^L_SR = I_{3\times 3}$ and $^S_WR(\phi_0) = I_{3\times3}$ for structural lines in vertical direction, we have a brief computation
\begin{equation}
{^Lm} = {^W_CR}\,\,K^{-1}m.
\end{equation}
The angular parameter $\theta$ is therefore determined by the horizontal heading in the local parameter space. We let $^Lm = (m_x,m_y,m_z)^T$. The angular parameter is computed as 
$\rev{\theta_0} = \text{atan2}(m_y, m_x)$, and the inverse depth $\rho$ is initialized as a preset $\rho_0$ for all newly detected structural lines. \rrev{We represent the initialization process as
\begin{equation}
\label{eq:l_0}
l_0 = \begin{bmatrix}
\theta_0\\
\rho_0
\end{bmatrix}=
\begin{bmatrix}
\Pi^{-1}(s,\phi_i,{^W_CR})\\
\rho_0
\end{bmatrix}
\end{equation}
The uncertainty of the initial parameters is set by the covariance
\begin{equation}
\label{eq:cov_0}
\Sigma_0 = \begin{bmatrix}
\sigma^2_{\theta_0} & 0\\
0 & \sigma^2_{\rho_0}
\end{bmatrix}
\end{equation}
where $\sigma_{\theta_0}$ is a small value that can be computed from back-projection $\Pi^{-1}(\cdot)$, with respect to the detection error of line segments ($2\sim 4$ pixels on average), the heading uncertainty of the Manhatton world and the orientation uncertainty of the current camera pose, both of which can be obtained from the filter. The uncertainty of inverse depth $\sigma_{\rho_0}$ is manually set to a large value $5$ to cover the distance from $0.2$ meter to $\infty $ meters.
}
\vspace{-0.8em}
\subsection{Triangulation of structural lines with prior knowledge}
\rev{Triangulation is called after each state update to renew the parameters for all lines. It is done by minimizing the sum of  squared re-projection errors (\ref{eq:reprojection_error}) among all views where the line is visible.} As we'll describe later,
the time interval of a line track usually exceed that of the historical views stored in the state. If we use only the observations in the visible views within the state, it usually leads to small motion parallaxes and produces inaccurate triangulation result.
If we use all visible views for triangulation, the computational cost may increase significantly and the obsolete pose estimates of the views outside of the state may also cause a large error.

 To address the mentioned problem, we maintain a prior distribution for each structural line, $\mathcal{N}(l_0,\Sigma_0)$, where
the mean value $l_{0} = (\theta_0, \rho_0)^T$ and the covariance matrix $\Sigma_{0} \in \mathbb{R}^{2\times2}$, to  store the initial prior distribution, or the prior distribution derived from the historical measurements as described later. The overall objective function is:
\begin{equation}
\label{eq:triangulation}
\small
\rrev{\mathop{\arg\min}_{l=(\theta,\rho)^T}}\sum_{k \in \mathcal{V}} r^2_k\rrev{(l)}/\sigma^2_{im} + (l - l_0)^T{\Sigma^{-1}_0}(l-l_0),
\end{equation}
where $r_k$ is the signed distances between the line segments and the projected lines defined in (\ref{eq:reprojection_error}) and $\mathcal{V}$ denotes the visible views in the state. $\sigma_{im}$ is a standard deviation describes the \rev{line detection} noise.  
\rrev{The mean $l_0$ and the covariance $\Sigma_0$ are initally set to (\ref{eq:l_0})(\ref{eq:cov_0}) and updated if some measurements in $\mathcal{V}$ are discarded as described later }.

This nonlinear least squares problem can be solved by Gauss-Newton algorithm in $3\sim5$ iterations\rrev{, even if the initial value is not close to the real one due to the high linearity introduced by inverse depth parameterization\cite{civera2008inverse}\cite{zhou2015structslam}.}

After triangulation, in order to track lines more reliably as described in Section \ref{sec:feature_detection}, 
we also update the range of structural lines $r = (r_{s}, r_{e})^T$ by intersection of this structural line with
the back-projection rays from two end points of the line segment $s$ in the last visible view.

 \subsection{\rrev{Handling dropped measurements of long line tracks}}\label{sec:marg}
Similar to sliding window estimators, one problem of our estimator is that features can be tracked in a period of time longer than that of views stored in the state. In existing sliding window estimators, those measurements outside of the sliding window are simply discarded in both key-frame based \cite{leutenegger2015keyframe} and filter-based \cite{mourikis2007multi} frameworks. This could lead to inaccurate estimates of line parameters as measurements outside of the sliding window still carries rich information about the line's geometry. In \cite{li2013optimization}, authors put those features that are tracked longer than the sliding window  into the state of filter. This is similar to classic EKF-SLAM framework \cite{davison2007monoslam} - the disadvantage is that the number of points put into the state needs be strictly controlled so that the state dimension will not become too high.

 \rrev{We propose here a novel method to convert those dropped measurements of a long track into a prior distribution about the line geometry to facilitate future update. We call this process as \emph{information accumulation}. This approach can be also applied to point features. We describe here the details for lines only.

  Let $\mathcal{D}$ be the set of poses being removed from the state (or the sliding window), and let the mean and covariance of the old prior distribution be $l^{old}_0$ and $\Sigma^{old}_0$. After
  $\mathcal{D}$ frames being removed from the sliding window, the prior distribution \rrev{needs to be updated to incorporate the information of dropped measurements on those frames.} The new mean $l^{new}_0$ is computed
by minimizing the objective function:
  \begin{equation}
\label{eq:marginalization}
  \small
\rrev{\mathop{\arg\min}_{l=(\theta,\rho)^T}}  \sum_{k \in \mathcal{D}} r^2_k\rrev{(l)}/\sigma^2_{im} + (l -l^{old}_0)^T{(\Sigma^{old}_0)^{-1}}(l-l^{old}_0),
  \end{equation}
  which is also minimized by Gauss-Newton algorithm.

Different from (\ref{eq:triangulation}), the minimization of (\ref{eq:marginalization}) is conducted only on dropped measurements. 
Let 
$\Lambda \delta l = Y$ be the normal equation being solved in the last Gauss-Newton iteration. The new covariance is \rrev{updated} as
  $\Sigma^{new}_0 \leftarrow \Lambda^{-1}$ before calling the next triangulation (\ref{eq:triangulation})}.

\begin{figure}
 \centering
\includegraphics[width=0.6\linewidth]{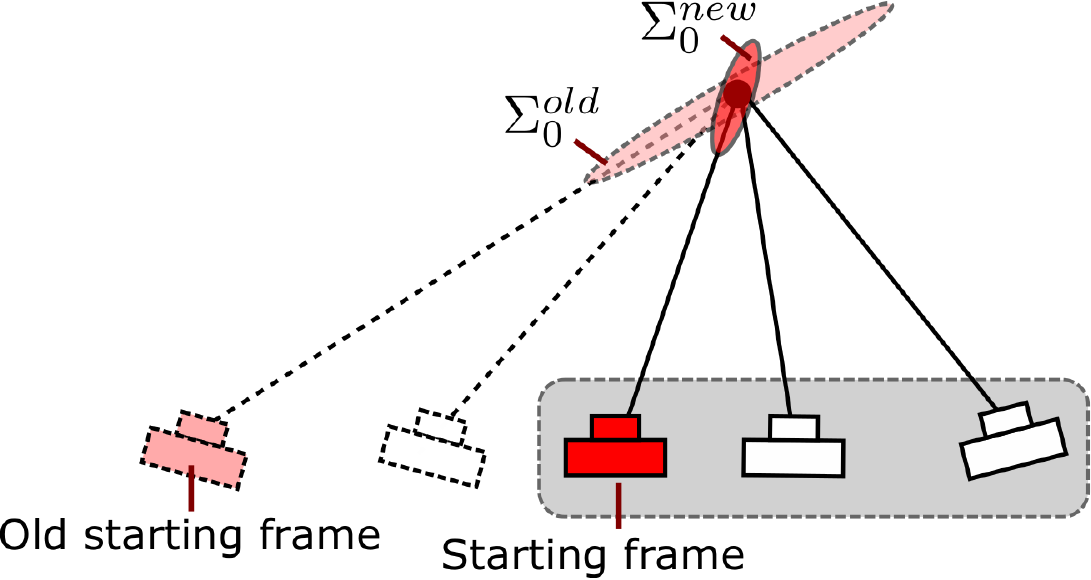}
\caption{Illustration of information accumulation. After the old poses (dashed)  are removed from the state, the covariance of prior distribution is adjusted ($\Sigma^{old}_0 \rightarrow \Sigma^{new}_0$) to incorporate the information from the removed measurements. The active poses included in the state are marked by the gray shadow. }
 \label{fig:marginalization}
 \end{figure}

Note that each structural line is anchored to the one of the camera poses (starting frame) in the state vector. If the starting frame is about to be removed from the state vector,
we need to firstly change the starting frame, $\{S\}$, to one of the remaining poses in the state, $\{S'\}$. Let ${^{S'}_S\mathcal{T}}$ be the transformation of the line parameter from the old starting frame into the new starting frame and $J$ be the Jacobian matrix of the coordinate transformation function ${^{S'}_S\mathcal{T}}(\cdot)$. We have $l' \leftarrow {^{S'}_S\mathcal{T}}(l), l'_0 \leftarrow {^{S‘}_S\mathcal{T}}(l_0)$. The covariance
matrices are updated $\Sigma' \leftarrow J \Sigma J^T, \Sigma'_0 \leftarrow J \Sigma_0 J^T$. The process of information accumulation could be better understood in Figure \ref{fig:marginalization}.

As shown in the experiments (Section \ref{sec:exp_marg}), the RMSE error will reduce to $60\%$ of the original one if we adopt the information accumulation in our VIO implementation.

 \subsection{Detecting and Merging Manhattan worlds}
 Detection of a Manhattan world in the image involves identifying vanishing points by clustering parallel lines into different groups \cite{toldo2008robust}. The vanishing points from those parallel groups are then extracted to determine the orientation of three axes in 3D space. The process  however becomes much simpler if an IMU is available,  since the accelerometer renders the vertical direction observable because of gravity. We adopt a similar approach \cite{camposeco2015using} to detect new Manhattan worlds in the image.  We start Manhattan world detection whenever vertical lines \rev{have been} identified as described in Section \ref{sec:line_classification}.  The vanishing line of the ground plane ($XY$ plane in the world frame) is computed as $\rrev{{l}_\infty} =
K^{-T} \, {^C_WR} \,[0,\,0,\,1]^T.$

 After that, we run an $1$-line RANSAC algorithm to detect possible Manhattan worlds as the following steps:

 1) randomly select a line segment that has not been identified as a structural line and extend it to intersect the horizontal vanishing line $l_{\infty}$ with a vanishing point $v_x$,  about which we make the assumption that it is the projection of $X$ direction of the possible Manhattan world. Since the vertical direction is already known, we are able to get direction of the Manhattan world $\phi$ and the vanishing point of $Y$ direction $v_{y}$.

 2) get the number of consistent line segments with the two vanishing points $v_x$ and $v_y$ in a similar approach as described in Section \ref{sec:line_classification}.

 3) repeat the above steps until the maximum number of iterations arrives.

 Finally, the cluster with the largest number of consistent line segments is considered as a possible Manhattan world. We further check if the number of consistent line segments
 is larger than a threshold ($4$ in our system) and larger than the number of existing structural lines (horizontal lines only) in the image.

Let $\phi^*$ be the orientation of the Manhattan world
 under detection. It also requires not to be close to any orientations of existing Manhattan worlds, namely $|\phi^* - \phi_i| > \Delta \phi, \forall i$, where $\Delta \phi$ is set to be $5^{\circ}$ in our implementation.
 Once all the conditions are satisfied, the detected Manhattan world $\phi^*$ is added into the state, and covariance is updated as described in Section \ref{sec:structure_line}.

Sometimes the orientation difference between two Manhattan world may be smaller than $\Delta \phi$ after a serials of EKF updates. In that case,  we merge the  two Manhattan worlds by removing the newer one from the state and adjust
the covariance accordingly. \rev{Structural lines anchored to the removed Manhattan world are also moved to the remaining one}.

 \subsection{Outlier rejection}
Outliers are detected by a Chi-squared gating test before EKF update.
According to the the measurement equation (\ref{eq:null_space_measurement}), the test is done by checking
\begin{equation}
\small
(z^{(0)})^T ( H^{(0)}P_{k|k-1}(H^{(0)})^T+\sigma^2_{im} I)^{-1} z^{(0)} < \chi_{0.95},
\end{equation}
where $\chi_{0.95}$ corresponds to the confidence level of $95 \%$.
Those structural lines without passing the gating test are excluded from EKF update. After EKF update, we re-triangulate all structural lines
 and further check the reprojection errors  (\ref{eq:reprojection_error}) at all visible views. The structural line with reprojection error larger than a threshold (about $4$ pixels in our system) is  discarded. Our two-phase outlier rejection makes our system more robust again outliers than using only chi-squared gating tests.
 \vspace{-1 ex}
   \section{Experimental results}
   
\subsection{Benchmark tests}
We first evaluate the proposed method on the Euroc dataset \cite{burri2016euroc}. This dataset is collected by a visual-inertial sensor mounted on a quadrotor flying in three different environments, which are classified into \emph{machine hall} and \emph{VICON rooms}. In the machine hall, the ground truth trajectories were obtained by a laser tracker, while in the VICON rooms, the motion capture system is used to get the ground truth trajectories.

We name our method as \emph{StructVIO} and compare it with two state-of-the-art VIO methods: OKVIS \cite{leutenegger2015keyframe} and VINS \cite{qin2017vins}. Both OKVIS and VINS use only point features and adopt the optimization framework to estimate the ego motion. We use the default parameters provided by their authors  throughout the tests. We disable the loop closing in VINS to test only the odometry performance. For StructVIO, we set the maximum number of features points as $125$ and the maximum number lines as $30$. All parameters were kept fixed during all benchmark tests.
We use the RMSE and maximum error to measure the performance of VIO.

The benchmark scenes are relatively small and full of textures. In such small scenes that are cluttered and contain rich textures, the point-only approaches should work well. Nevertheless, we observe that exploring structural regularity still helps.

\begin{figure}
 \centering
 \includegraphics[width=0.99\linewidth]{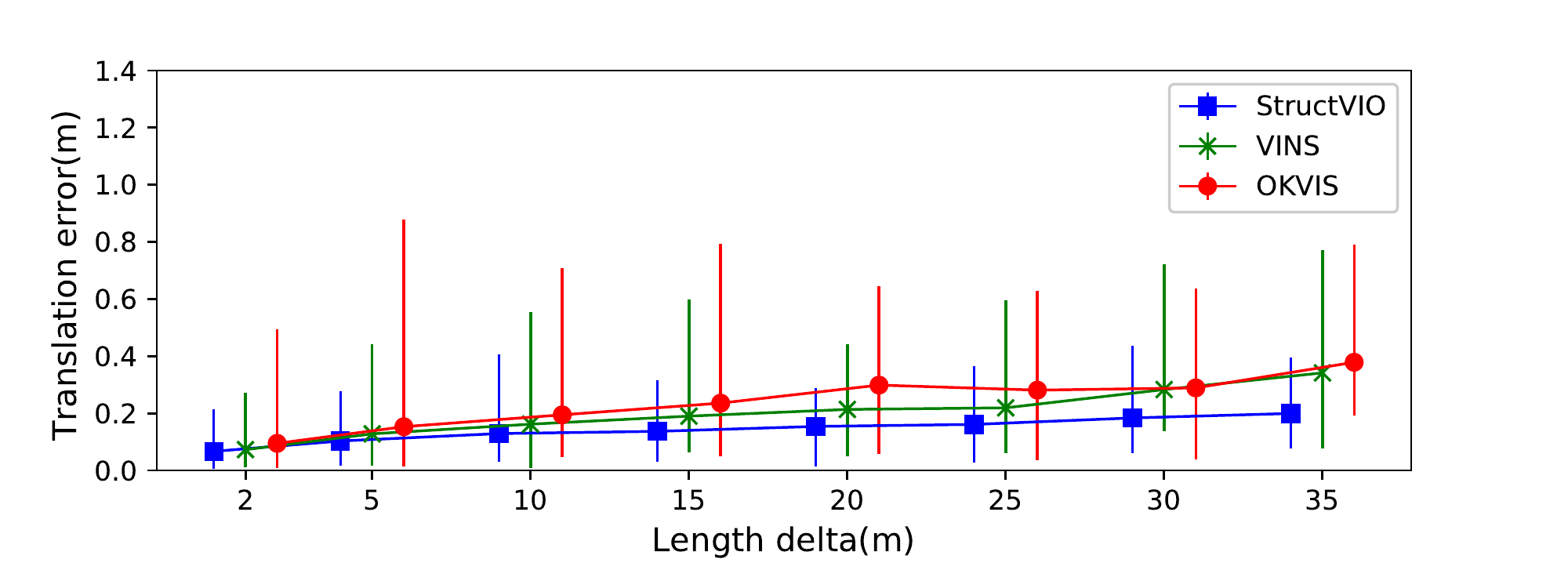}
 \caption{Relative position errors on EuRoC tests. The average error, minimum error and maximum error of all tests are shown.}
 \label{fig:euroc_rpe}
 \end{figure}

\begin{figure}[h]
 \centering
 \subfigure[Machine hall]{
 \label{fig:machine_hall_snapshot}
 \includegraphics[width=0.45\linewidth]{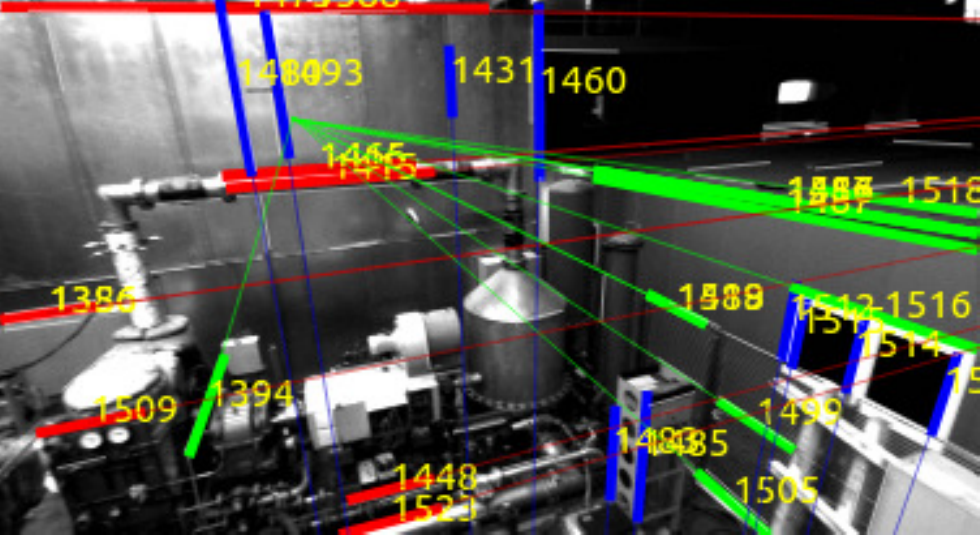}
 \includegraphics[width=0.45\linewidth]{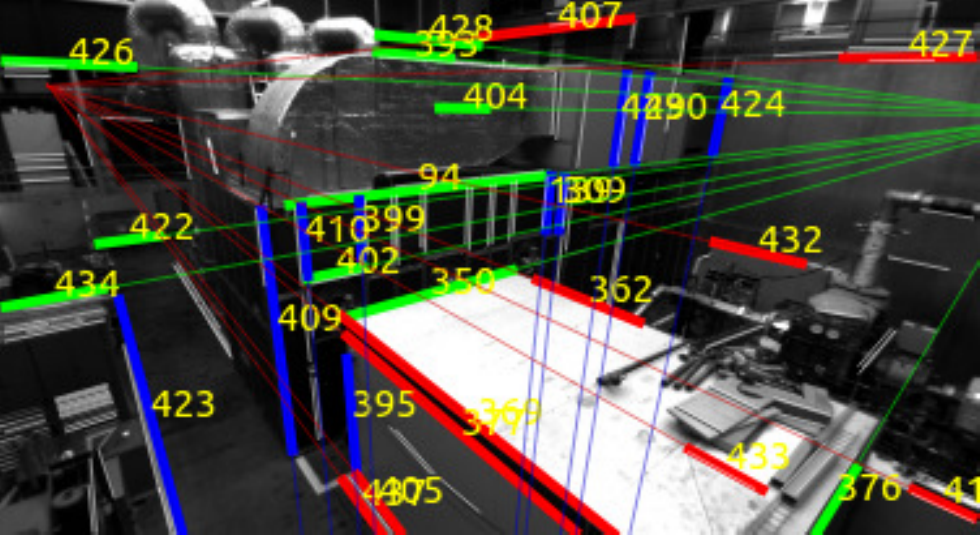}
 }

 \subfigure[Vicon room]{
 \label{fig:vicon_room_snapshot}
 \includegraphics[width=0.45\linewidth]{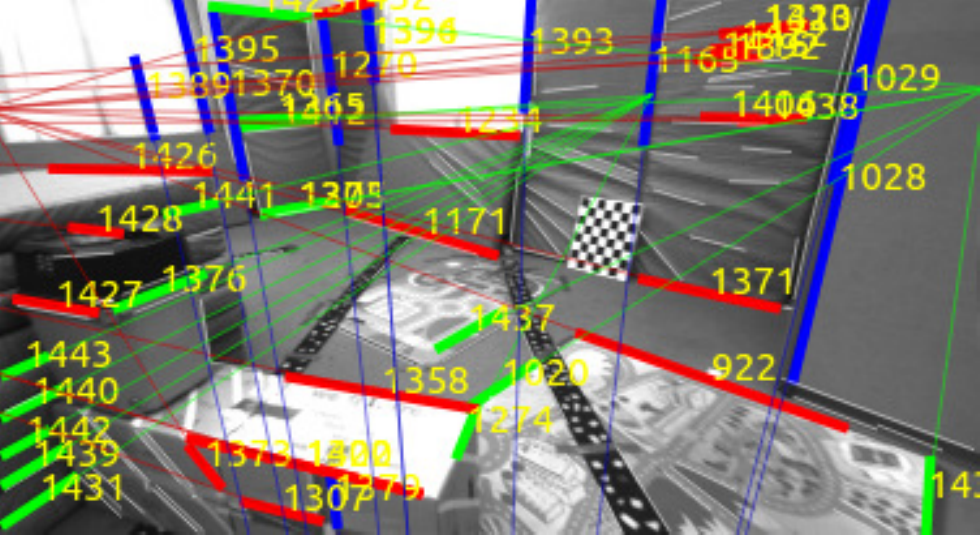}
 \includegraphics[width=0.45\linewidth]{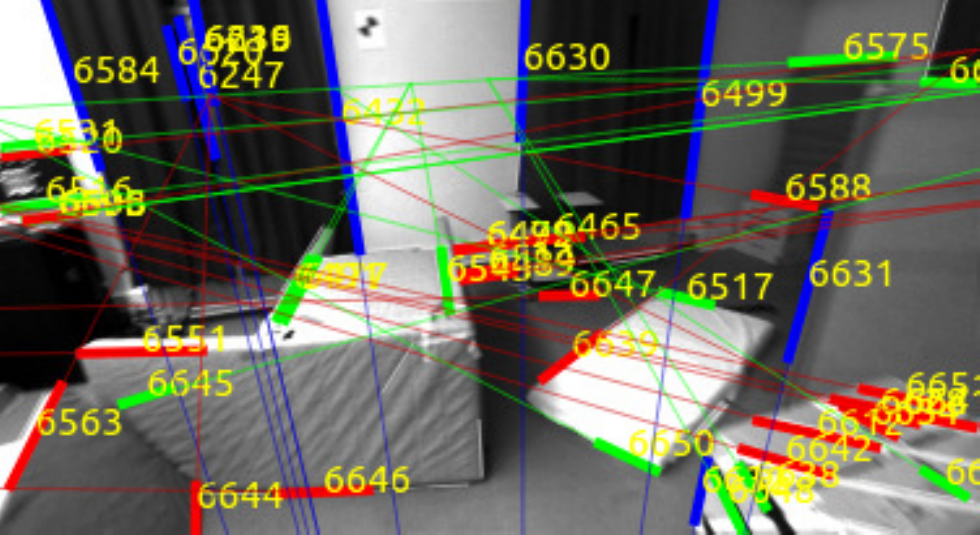}
 }
 \caption{Line features extracted in different scenes. Blue lines are classified as vertical. Red and green lines are classified as horizontal while not necessary in the same Manhattan world. In the machine hall, the majority of lines are aligned with three orthogonal directions, which are well described by a Manhattan world. In the Vicon room, the line features are more cluttered, so that multiple Manhattan worlds are used for parameterization of lines.}
\end{figure}

As shown in Table \ref{tab:euroc}, StructVIO performs better than the state-of-the-art VIO methods on all the sequences except \emph{V01\_02\_Medium}, where StructVIO's RMSE is slightly larger than VINS's.  \rrev{StructVIO also produces the lowest relative positional errors among all these methods as shown in Figure \ref{fig:euroc_rpe}.}
StructVIO correctly find the Manhattan regularity in the machine hall as shown in Figure \ref{fig:machine_hall_snapshot}. 
In the VICON room, we observe that our system reports multiple Manhattan worlds as shown in Figure \ref{fig:vicon_room_snapshot} due to cluttered line features on the ground, but they still help as those horizontal lines still encode heading constraints to reduce the overall drift.
Even if sometimes no Manhattan world has been detected, vertical lines still help since they immediately reveal the gravity direction to improve the pose estimates  for a moving IMU.

We also did quantitative analysis on the performance of using different number of features with different combinations of features, denoted by \emph{point-only}, \emph{point+line}, and \emph{StructVIO}, within the same filter framework adopted in this work. Here \emph{point+line} represents the VIO method that uses the combination of point and line features, where the line features are the general lines without the prior  orientation.

We repeatedly perform VIO on the same sequence by changing the maximum number of features from $50$ to $250$, and obtain the average RMSEs (and standard deviations) for different numbers. During the tests, we kept the maximum number of line features as $30$ and change the number of points for the methods that involve lines.

\begin{figure}[h]
 \centerline{
 \subfigure[Machine hall]{
 \includegraphics[width=0.5\linewidth]{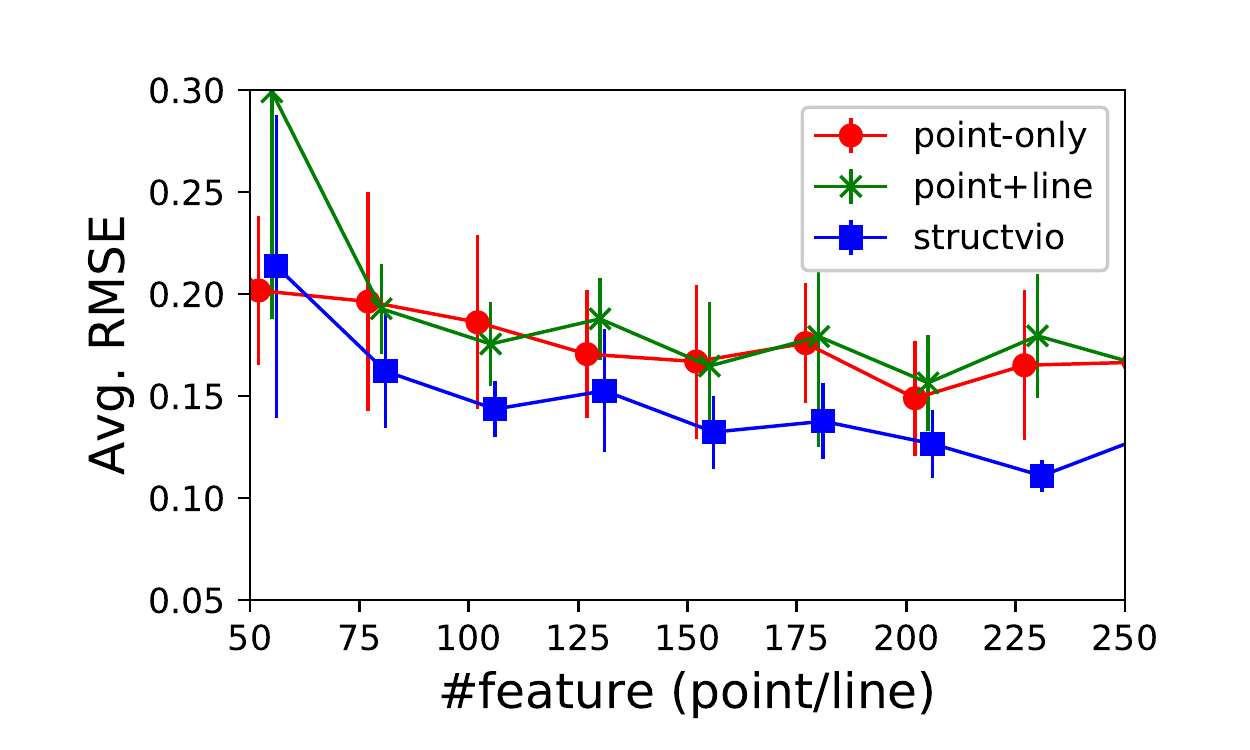}
 \label{fig:machine_hall_res}
 }
 \subfigure[Vicon room]{
 \includegraphics[width=0.5\linewidth]{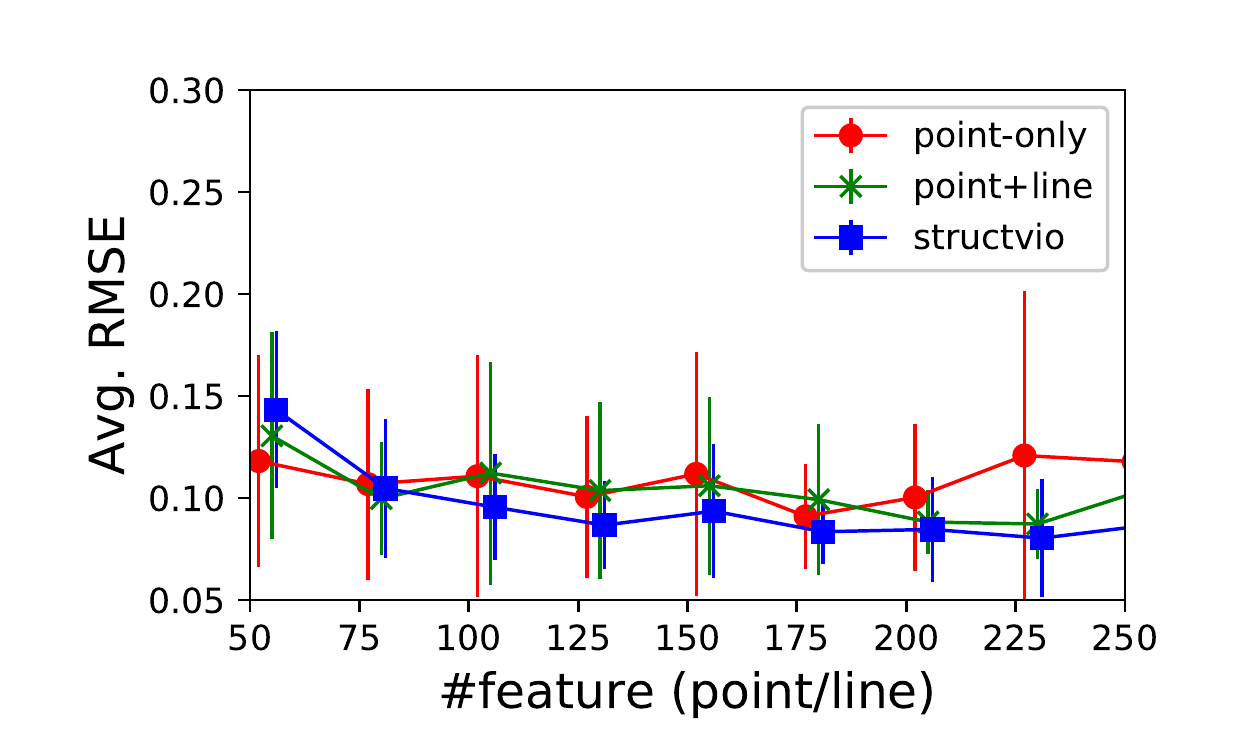}
 \label{fig:vicon_room_res}
 }
 }
 \caption{StructVIO performs better if the scene exhibit structural regularity. In (a), since the machine hall is more like a box world, StructVIO here performs significantly better. Note that the point+line method performs very close to the point-only method. In (b), since the vicon room is rich textured and heavily cluttered, the performance of different methods are close to each other.}
\end{figure}

To better understand how the structural information help, we plot the average RMSEs separately for the machine hall and the VICON room as shown in Figure \ref{fig:machine_hall_res} and \ref{fig:vicon_room_res}. As we can see, since the machine hall exhibits stronger regularity in structures and contains less textures, StructVIO leads to less RMSE than the point-only and point+line methods if the same number of features are used. In contrast, the VICON rooms are highly cluttered and full of textures, where all the methods have very close performances as shown in Figure \ref{fig:vicon_room_res}.

From these results, we may roughly conclude that structural information helps more in environments with strong regularities and few textures. As the scenes in the Euroc dataset are relatively small, this conclusion requires to be further tested. In the next section, we will conduct experiments in large indoor scenes.
\begin{table}[h]
\caption{Absolute position errors (in meters) of Euroc Benchmark tests. Both \emph{Rooted Mean Squared Error}(RMSE) and \emph{Maximum Error} (Max.) are presented. }
\begin{center}
  \resizebox{0.85\textwidth}{!}{ \begin{minipage}{\textwidth}  \small
\begin{tabular}{lrrrrrrrrrrrr}
\toprule
\multirow{2}{*}{Dataset}&  \multicolumn{2}{c}{OKVIS\cite{leutenegger2015keyframe}}& \multicolumn{2}{c}{  VINS\cite{qin2017vins}(w/o loop)}  &  \multicolumn{2}{c}{ StructVIO} \\
  \cmidrule{2-7}
         &  RMSE &  Max. &  RMSE &  Max. &  RMSE &  Max.  \\
\midrule
MH\_01\_easy	 &	0.308	 &	0.597	 &	\textbf{0.157}$^2$	 &	0.349	 &	\textbf{0.079}$^1$	 &	0.251\\ 
MH\_02\_easy	 &	0.407	 &	0.811	 &	\textbf{0.181}$^2$	 &	0.533	 &	\textbf{0.145}$^1$	 &	0.267\\ 
MH\_03\_medium	 &	0.241	 &	0.411	 &	\textbf{0.196}$^2$	 &	0.450	 &	\textbf{0.103}$^1$	 &	0.271\\ 
MH\_04\_difficult	 &	0.363	 &	0.641	 &	\textbf{0.345}$^2$	 &	0.475	 &	\textbf{0.130}$^1$	 &	0.286\\ 
MH\_05\_difficult	 &	0.439	 &	0.751	 &	\textbf{0.303}$^2$	 &	0.434	 &	\textbf{0.182}$^1$	 &	0.358\\ 
V1\_01\_easy	 &	\textbf{0.076}$^2$	 &	0.224	 &	0.090	 &	0.201	 &	\textbf{0.060}$^1$	 &	0.180\\ 
V1\_02\_medium	 &	0.141	 &	0.254	 &	\textbf{0.098}$^1$	 &	0.334	 &	\textbf{0.130}$^2$	 &	0.260\\ 
V1\_03\_difficult	 &	0.240	 &	0.492	 &	\textbf{0.183}$^2$	 &	0.376	 &	\textbf{0.090}$^1$	 &	0.263\\ 
V2\_01\_easy	 &	0.134	 &	0.308	 &	\textbf{0.080}$^2$	 &	0.232	 &	\textbf{0.045}$^1$	 &	0.140\\ 
V2\_02\_medium	 &	0.187	 &	0.407	 &	\textbf{0.149}$^2$	 &	0.379	 &	\textbf{0.066}$^1$	 &	0.157\\ 
V2\_03\_difficult	 &	\textbf{0.255}$^2$	 &	0.606	 &	0.268	 &	0.627	 &	\textbf{0.110}$^1$	 &	0.231\\ 
\bottomrule
\end{tabular}
\end{minipage} }
\end{center}
\label{tab:euroc}
\end{table}

\subsection{Large Indoor Scenes}
In this section, we conduct experiments to test our method in large indoor scenes.
We use the Project Tango tablet to collect image and IMU data for evaluation. Gray images are recorded at about  $15$ Hz and IMU data at $100$ Hz. Data collection starts and ends with the same location,  while traveling in different routes.  
We use Kalibr \cite{furgale2013unified} to calibrate the IMU and camera parameters.  To run our algorithm, we remove the distortion and extract the line features from the distortion-free images, while we extract the point features from the original images.

\begin{figure}
  \centering
  \includegraphics[width=0.32\textwidth]{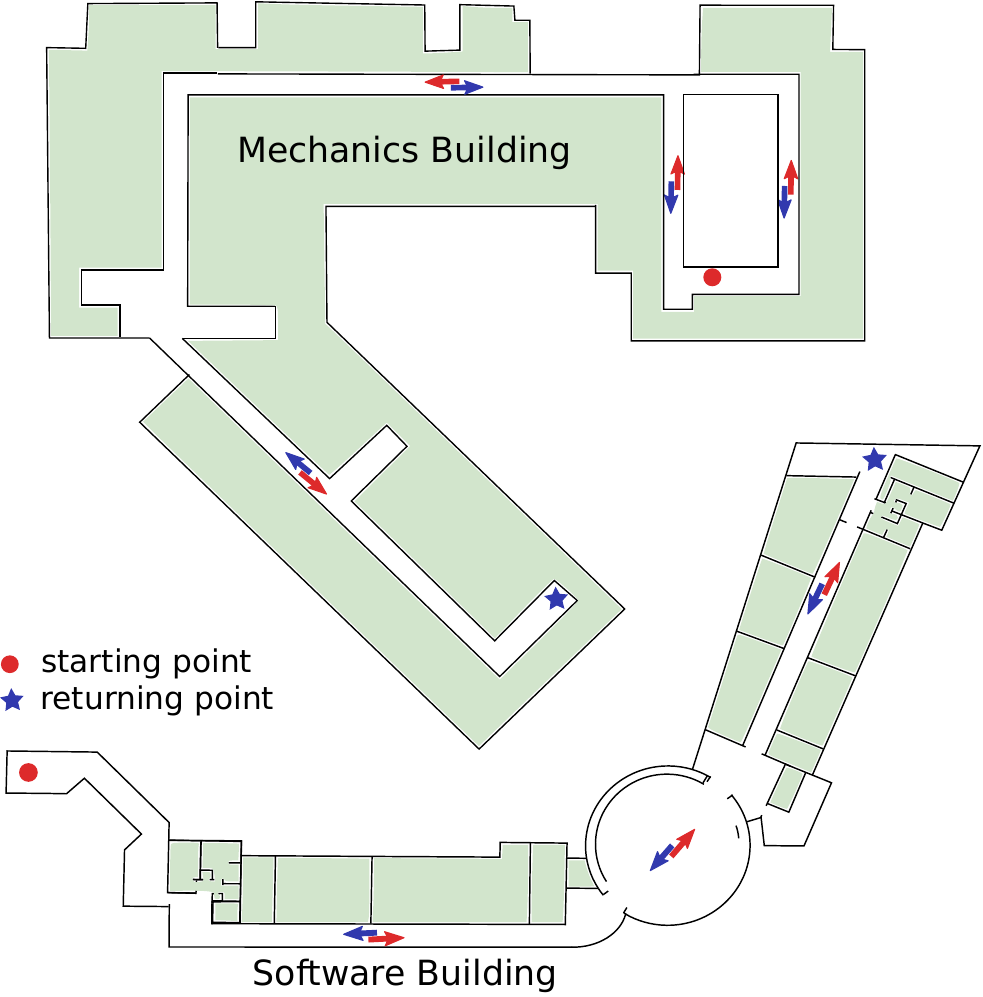}
  \caption{Two buildings where we collected our data, which are denoted by \emph{Mech} and \emph{Soft} respectively in the text. Note that both buildings have irregular structures such as oblique corridors and circular halls other than a single box world. }
  \label{fig:floor_plans}
 \end{figure}

 Data collection were conducted within three different buildings, where the camera experiences  rapid camera rotation, significant change of lighting conditions, distant features,  and lack of textures\footnote{The StructVIO executable and datasets can be downloaded from \emph{https://github.com/danping/structvio}}. The buildings are referred to as \emph{Micro}, \emph{Soft}, and \emph{Mech} respectively.The \emph{Micro} building well fits the Manhattan world assumption since it has only three orthogonal directions, while the \emph{Mech} and \emph{Soft} buildings have oblique corridors and curvy structures that can not be modeled by a single box world as shown in Figure \ref{fig:floor_plans}.
 
\begin{figure*}
 \centering
   \includegraphics[width=0.95\linewidth]{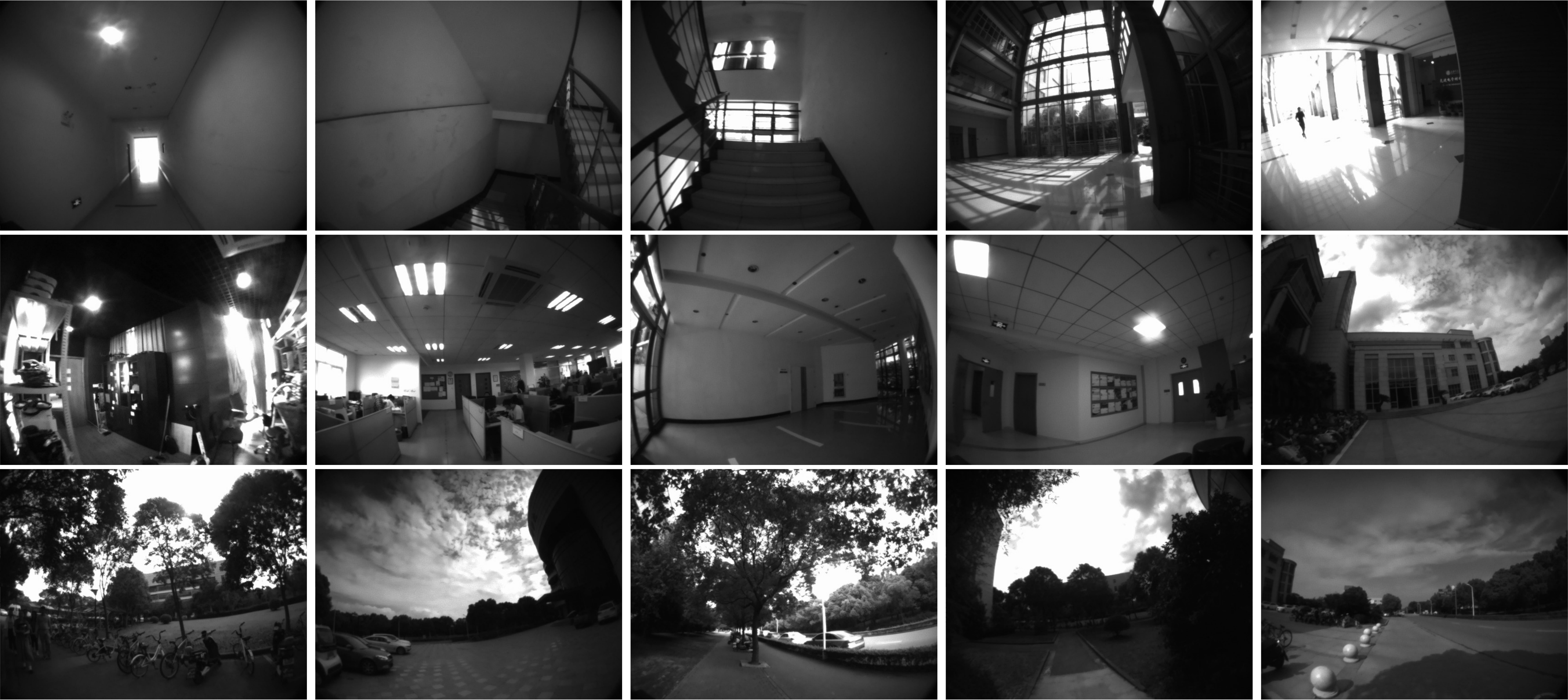}
\caption{Our datasets for evaluation of visual-inertial odometry methods. Datasets are collected inside and outside of three buildings. The indoor parts include typical scenes such as narrow passages, staircases, large halls, clutter workshop, open offices, corridor junctions and so on. The outdoor parts include trees, roads, parking lots, and building entrance. Challenging cases such as over or under exposure, texture-less walls, distance features, and fast camera motions can be found in our datasets. }
 \label{fig:datasets}
 \end{figure*}

 During data collection, the collector also went out of the building and walked around from time to time. Each data collection lasts about $5 \sim 10$ minutes, and the traveling distances are usually several hundreds meters. Some of the captured images are shown in Figure \ref{fig:datasets}. Our datasets exhibit many challenging cases, including over/under exposure of images, distance features, quick camera rotations and texture-less walls.
 
 \paragraph{Evaluation method}
 \begin{figure}[h]
 \centering
 \includegraphics[width=0.8\linewidth]{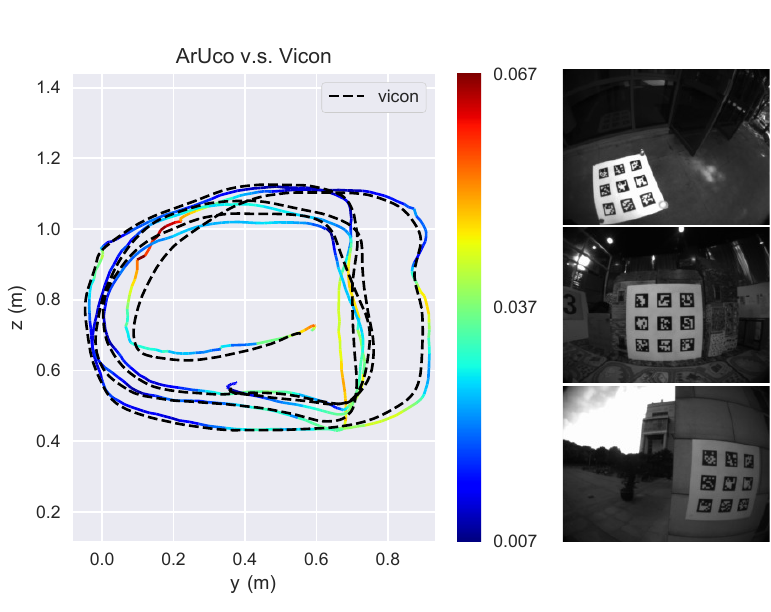}
\caption{ArUco markers are used for estimating the end-to-end positional errors. 
As shown on the left, the camera trajectory estimated from ArUco marker is very close to the Vicon ground truth, where the positional difference is about $2.8cm$ on average. We therefore use the camera trajectory 
estimated from the ArUco marker as the reference in our long-travel tests.}
 \label{fig:aruco}
 \end{figure}
 
Unlike in a small indoor environment, we cannot use motion capture systems to get the ground truth of the whole trajectory in such large indoor scenes to exactly evaluate the performance. Instead, we managed to obtain the ground truth trajectory in the beginning and in the end. The first way is that we start the data collection from a vicon room and return to it when finished. The second way is to use a printed QR pattern (ArUco marker \cite{de2015aruco}) to get the camera trajectory when there is no motion capture system available. Though using the ArUco marker leads to less accurate result, it still produces trajectories with about $2.8$ cm accuracy (validated by the vicon system as shown in Figure \ref{fig:aruco}), which is sufficient to treated as the ground truth in our tests.
 
We aligned the estimated trajectory with the ground truth trajectory acquired in the beginning and compared the difference between the estimated trajectory and the ground truth trajectories( acquired both at the beginning and at the end of data collection). The different is described by RMSE and Max. of absolute pose error after alignment. 

Let $p^t$ be the camera trajectory estimated by the VIO algorithm and $g^t_s$, $g^t_e$ be the ground truth trajectories estimated at the beginning and at the end respectively, from either VICON or ArUco. First we obtain the transformation $T$ that
\begin{equation}
 T^* = \arg\min_T \sum_{t \in s} (\|T(p^t) - g^t_s\|^2).
\end{equation}
The RMSE and Max. are computed as 
\begin{equation}
\sqrt{\frac{1}{|e|}\sum_{t \in e} \|(T(p^t)-g^t_e \|^2)},\text{and}\,\max{|T(p^t)-g^t_e|}.
\end{equation}

Note that the two errors in fact does not fully describe the performance of a visual-inertial system. It happens occasionally that a visually bad trajectory (intermediate pose estimates are not good) ends up with a nearly closed loop result. We therefore test repeatedly in one scene in order to reduce biases. 


\begin{figure*}[h]
 \centering
 \includegraphics[width=0.8\linewidth]{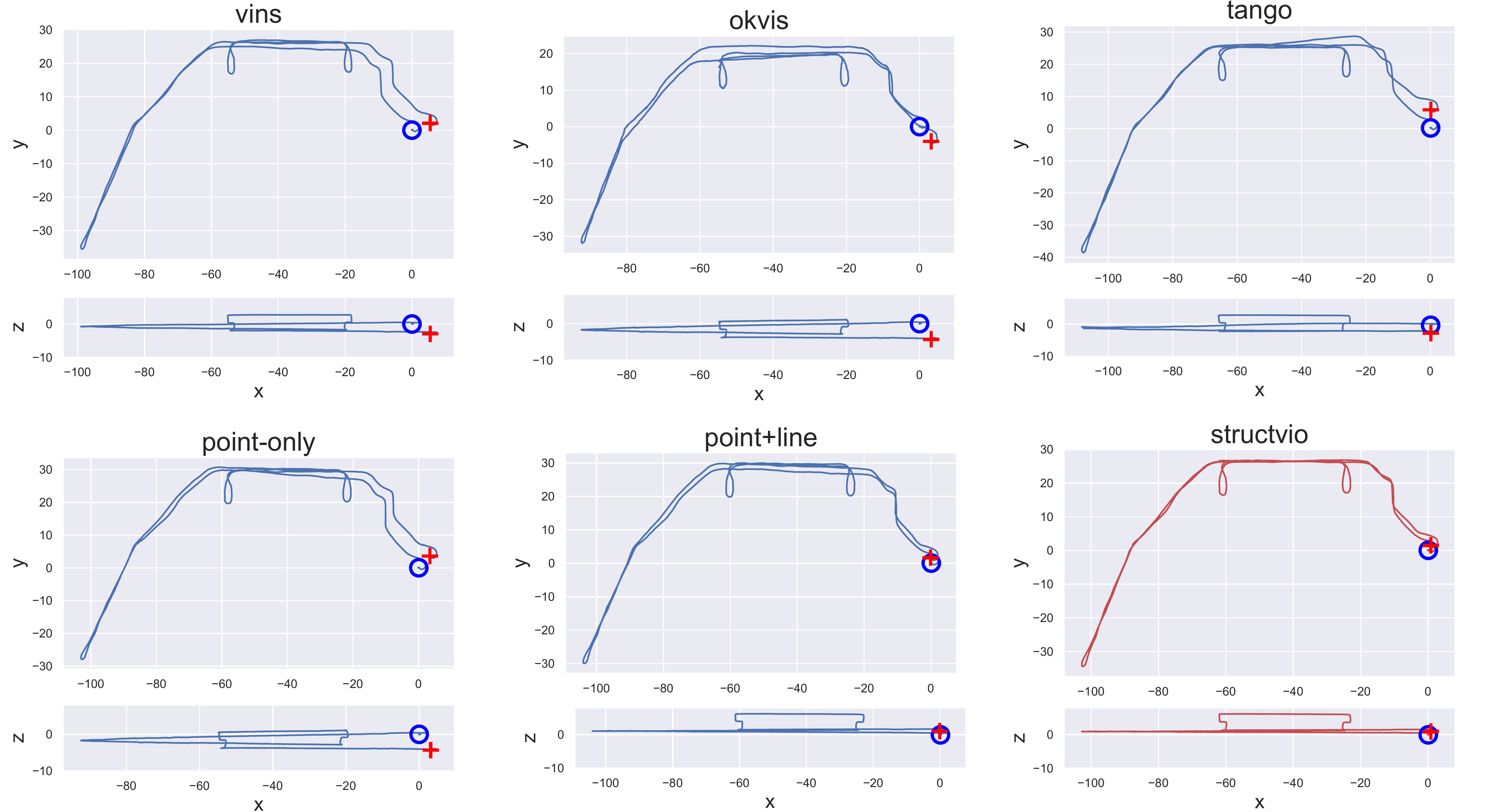}
 \caption{Results of \emph{Soft-02} sequence. In this case, the camera travels in the \emph{Software building} firstly on the first floor and then on the second floor. The traveling distance is about $438$ meters. Both trajectories on $xy$ and $xz$ planes are shown. It clearly shows that the proposed method (structvio) produces the best accuracy in this case (about $0.33\%$ drift error). We can note other methods produces significant drifts in both horizontal and vertical directions. Please look at Table \ref{tab:tango_results} for RMSE errors of other methods. }
 \label{fig:soft02}
\end{figure*}

\begin{figure*}[h]
 \centering
 \includegraphics[width=0.8\linewidth]{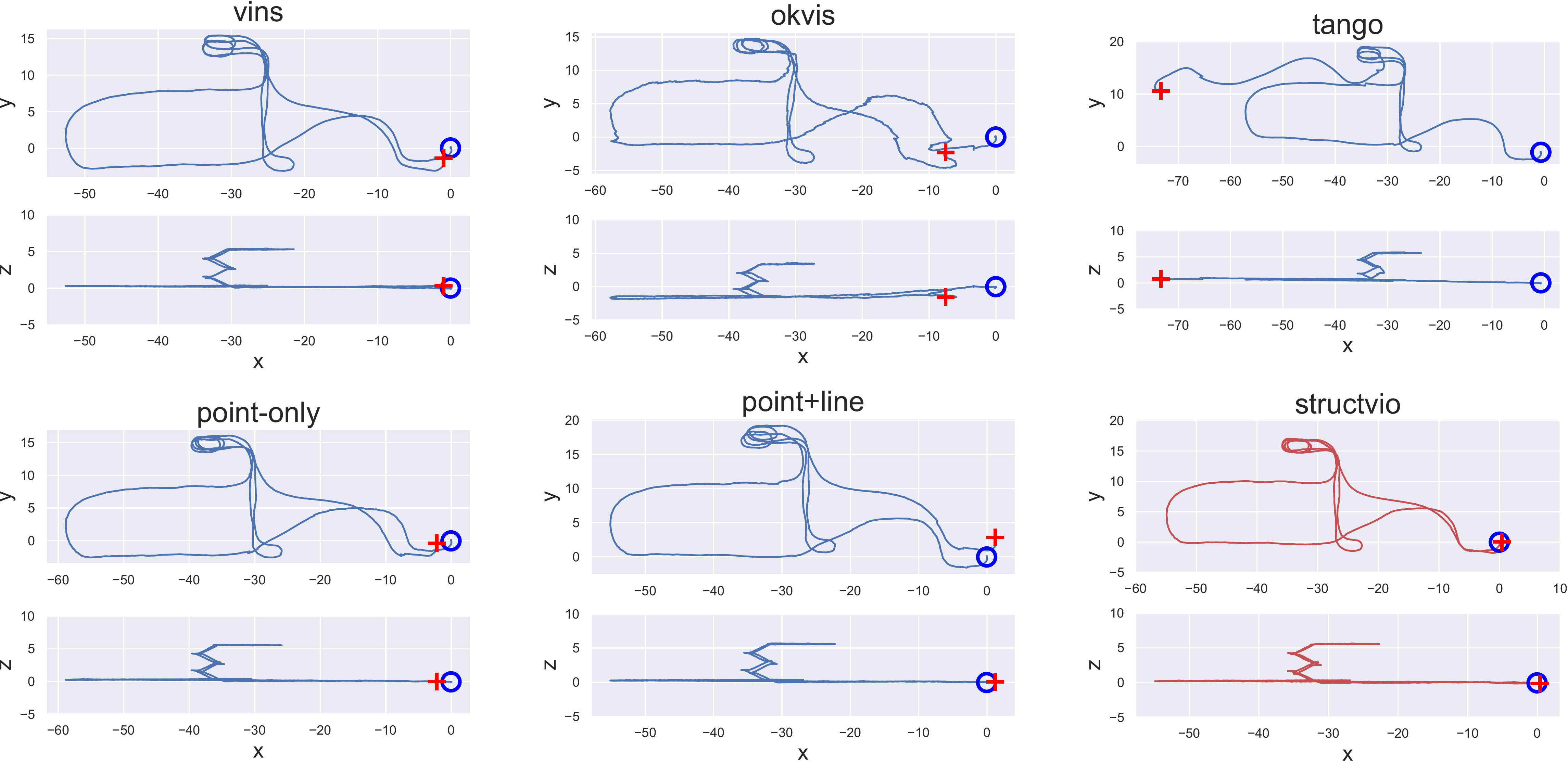}
 \caption{Results of \emph{MicroA-04}. In this case, the data collector started from the outside of the \emph{Micro-Electronic Building},  walked into this building, went up the stairs  and returned back to the starting point. The traveling distance is about $237.9$ meters. Since the staircase mainly contains texture-less walls and lacks of sufficient illumination, we observe that the VIO system inside of Tango produced a large error when the camera was moving in the staircase. OKVIS does not perform good after initialization. This may be caused by the change of illumination when the camera entered the building. Compared with VINS, our structvio method performs much better. This is better perceived if we check the trajectories at the area of staircase.}
 \label{fig:microa04}
\end{figure*}


\begin{figure*}
 \centering
 \includegraphics[width=0.8\linewidth]{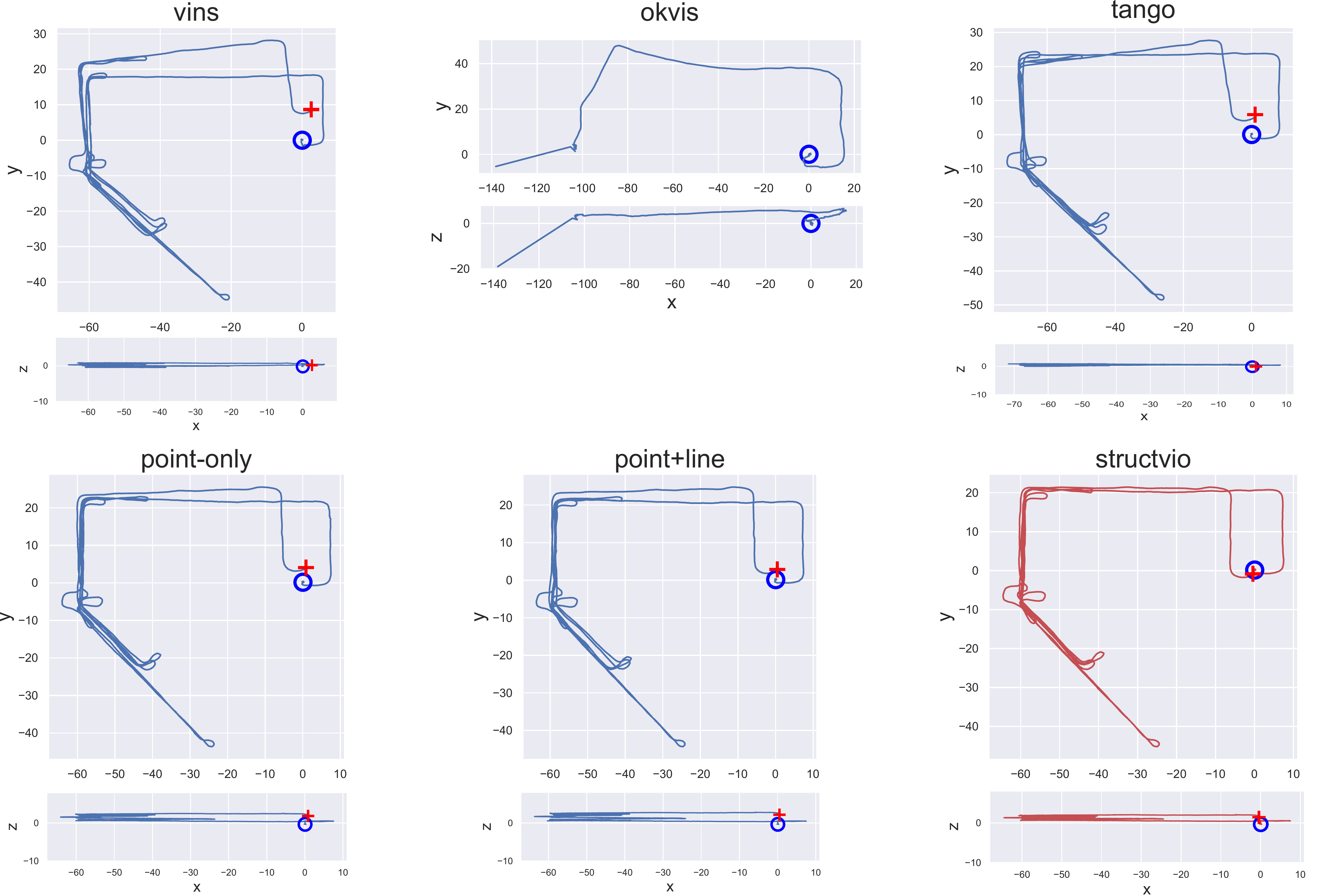}
 \caption{Results of \emph{Mech-04}. The data collector walked around on the same floor inside of \emph{Mechanical Building}, where the traveling distance is about $650$ meters. Note that a lot of turns were conducted during data collection. OKVIS failed after the third turns. VINS and Tango performed similarly and produced large orientation errors. Our point-only and point+line methods performed better but produced large orientation errors as well. The proposed structvio method, however, estimated the orientation correctly throughout the sequence, which in turn led to a small drift error $0.11\%$.}
 \label{fig:mech01}
\end{figure*}

\paragraph{Experiments}
\begin{table*}[t]
\caption{\rrev{Loop closing error} of different algorithms in three indoor scenes: RMSE error (pos. [$m$]) max error (drift [$m$]) . The traveling distance is approximated by the trajectory length produced by StructVIO.  The bar `-' indicates the algorithm fails to finish the whole trajectory. The first and second best results are in black and notated by their ranks.
}
    \resizebox{0.82\textwidth}{!}{ \begin{minipage}{\textwidth}  \small
   \begin{tabular}{lccccccccccccc}
    \toprule
     \multirow{2}{*}{Seq. } &\multirow{2}{*}{  Traveling } & \multicolumn{2}{c}{OKVIS\cite{leutenegger2015keyframe}} & \multicolumn{2}{c}{VINS\cite{qin2017vins}(w/o loop)} &
    \multicolumn{2}{c}{Project Tango} & \multicolumn{2}{c}{Point-only} &\multicolumn{2}{c}{Point+Line} & \multicolumn{2}{c}{StructVIO}\\
     \cmidrule(r){3-4}  \cmidrule(r){5-6}  \cmidrule(r){7-8}   \cmidrule(r){9-10} \cmidrule(r){11-12} \cmidrule(r){13-14}
     \multicolumn{1}{l}{Name}& Dist. [$m$] &
     \multicolumn{1}{l}{RMSE}  & \multicolumn{1}{l}{Max.} &
   \multicolumn{1}{l}{RMSE }  & \multicolumn{1}{l}{Max.} &
   \multicolumn{1}{l}{RMSE }  & \multicolumn{1}{l}{Max. } &
   \multicolumn{1}{l}{RMSE}  & \multicolumn{1}{l}{Max. } &
   \multicolumn{1}{l}{RMSE}  & \multicolumn{1}{l}{Max. } &
   \multicolumn{1}{l}{RMSE}  & \multicolumn{1}{l}{Max} \\
\midrule
Soft-01	&315.167	&6.702	&9.619	&4.861	&6.688	&5.715	&8.181	&\textbf{2.153}$^2$	&2.728	&2.262	&2.842	&\textbf{1.931}$^1$	&2.437\\
Soft-02	&438.198	&4.623	&6.713	&2.713	&4.086	&4.238	&6.226	&3.905	&5.243	&\textbf{1.468}$^2$	&2.026	&\textbf{1.429}$^1$	&1.984\\
Soft-03	&347.966	&\textbf{4.505}$^2$	&6.223	&7.270	&9.832	&167.825	&228.630	&6.515	&8.119	&8.618	&10.790	&\textbf{0.325}$^1$	&1.020\\
Soft-04	&400.356	&3.993	&5.784	&28.667	&75.479	&2.453	&3.544	&\textbf{1.550}$^1$	&2.028	&4.051	&5.262	&\textbf{1.722}$^2$	&2.241\\
\midrule
Mech-01	&340.578	&3.627	&4.745	&2.452	&3.260	&\textbf{1.948}$^2$	&2.726	&3.298	&3.961	&4.323	&5.181	&\textbf{0.909}$^1$	&1.165\\
Mech-02	&388.548	&3.079	&4.195	&3.570	&4.754	&\textbf{1.596}$^2$	&2.217	&1.663	&2.108	&2.317	&2.927	&\textbf{0.779}$^1$	&1.022\\
Mech-03	&317.974	&3.875	&5.324	&4.682	&9.113	&4.220	&5.781	&\textbf{2.384}$^2$	&3.020	&4.193	&5.272	&\textbf{1.161}$^1$	&1.532\\
Mech-04	&650.430	&-	&-	&3.002	&8.592	&1.915	&5.808	&1.785	&4.663	&\textbf{1.425}$^2$	&3.729	&\textbf{0.742}$^1$	&1.940\\
\midrule
MicroA-01	&257.586	&2.485	&3.382	&\textbf{0.654}$^2$	&1.148	&45.599	&61.058	&2.849	&3.505	&2.189	&2.721	&\textbf{0.642}$^1$	&1.225\\
MicroA-02	&190.203	&3.428	&5.186	&14.222	&57.172	&\textbf{1.145}$^1$	&1.692 &1.964	&2.514	&\textbf{1.723}$^2$	&2.207	&2.089	&2.661\\
MicroA-03	&388.730	&0.078	&0.779	&\textbf{1.800}$^1$	&2.578	&4.400	&6.253	&3.824	&5.169	&3.072	&4.232	&\textbf{1.884}$^2$	&2.892\\
MicroA-04	&237.856	&6.136	&8.532	&\textbf{0.994}$^2$	&1.765	&55.200	&75.318	&2.056	&2.897	&2.406	&2.879	&\textbf{0.350}$^1$	&0.448\\
\midrule
MicroB-01	&338.962	&2.898	&4.025	&\textbf{1.856}$^2$	&2.944	&38.197	&50.572	&7.084	&8.576	&7.337	&8.913	&\textbf{1.477}$^1$	&1.902\\
MicroB-02	&306.316	&2.240	&3.490	&\textbf{1.030}$^2$	&2.431	&5.660	&8.652	&2.521	&3.714	&3.197	&4.610	&\textbf{0.470}$^1$	&0.799\\
MicroB-03	&485.291	&-	&-	&2.132	&3.368	&\textbf{2.009}$^2$	&2.960	&6.490	&8.978	&4.507	&6.301	&\textbf{0.445}$^1$	&0.675\\
MicroB-04	&357.251	&4.064	&6.481	&\textbf{1.332}$^2$	&2.068	&13.962	&22.028	&5.078	&7.713	&1.977	&3.074	&\textbf{0.473}$^1$	&0.777\\
\midrule
Mean Drift Err.(\%) &  &1.078\% & &1.410\% & &6.180\% & &0.957\% & &\textbf{0.956\%}$^2$ & &\textbf{0.292\%}$^1$ &\\
Median Drift Err.(\%) &  &0.781\% & &\textbf{0.538\%}$^2$ & &0.900\% & &0.559\% & &0.570\% & &\textbf{0.176\%}$^1$ &\\
\bottomrule
    \end{tabular}
    \end{minipage} }
\label{tab:tango_results}
\end{table*}

We conduct experiments to test the performance of approaches using different combinations of features with/without structural constraints. The first approach (Point-only) uses only points. The second approach (Point+Line) uses both points and lines but without using the structural constraints. The last approach (StructVIO) uses both points and lines with the Atlanta world assumption. We keep the maximum number of points as $150$ and the maximum number of lines as $30$ during all tests.

We also present the results from the Tango system and the other two state-of-the-art VIO algorithms OKVIS\cite{leutenegger2015keyframe}  and VINS \cite{qin2017vins} for comparison. We use the default parameters and implementation by their authors and disable loop closing in VINS to test only the performance of odometry. 
We also add the FOV distortion model \cite{devernay2001straight}  to OKVIS and VINS softwares to enable them to process the raw image data from Tango, as Tango uses the FOV model to describe the camera distortion. Parameters are kept constant for all algorithms during tests.

\paragraph{Results}

The results are presented in Table \ref{tab:tango_results}. We listed the traveling distances of each sequence and the  positional errors of all algorithms. In the bottom of the table, we also compute the mean/median drift error as the average/median RMSE position error divided by the average traveling distance.

We can see that from the results our approach (StructVIO) using structural lines achieves the best performance among all methods in extensive tests in large indoor scenes. This is largely due to two reasons: 1) structural lines are a good complement to point features in low-textured scenes within in man-made environments; 2) structural lines encode the global orientations of the local environments, and render the horizontal heading observable. The limited heading error therefore reduces the overall position error. We can see that in the results, the mean drift error reduce from $0.956\%$ to $0.292\%$ with the help of structural lines under Atlanta world assumption.

However, if we use line features without considering if they are structural lines, the results (Point+Line) show that it does not improve the accuracy much in our tests. Note the average/median drift errors of the point-only approach and the point+line approach are almost the same ($0.957\%/0.559\%$ versus $0.956\%/0.570\%$) . Similar phenomenon has been observed in \cite{zhou2015structslam}. The reason could be that the general lines have more degree of freedom and are less distinguishable than points. Both facts can sometimes have negative impact on the VIO system as discussed in \cite{zhou2015structslam}.


\begin{table}[h]
\caption{Results of using Atlanta world and Manhattan world assumptions}
\centering
\begin{tabular}{lllll}
\toprule
\multirow{2}{*}{Seq. Name} & \multicolumn{2}{l}{Atlata world} & \multicolumn{2}{l}{Manhattan world} \\      \cmidrule(r){2-3}  \cmidrule(r){4-5}  
                  & RMSE            & Max.            & RMSE              & Max.             \\ \cmidrule(r){1-5}
Mech-01&0.909&1.165&1.144&1.524\\
Mech-02&0.779&1.022&1.286&1.061\\
Mech-03&1.161&1.532&2.029&1.211\\
Mech-04&0.742&1.940&1.822&2.193\\
\midrule
Soft-01&1.931&2.437&2.896&2.397\\
Soft-02&1.429&1.984&3.092&4.149\\
Soft-03&0.325&1.020&3.352&4.236\\
Soft-04&1.722&2.241&3.178&4.120\\
\bottomrule
\end{tabular} \label{tab:maha_vs_atlan}
\end{table}

Another interesting observation is that both optimization-based methods (OKVIS and VINS ) perform worse than the our
point-only approach and sometimes fail in our tests, though they are theoretically more accurate than the filter-based approach. The first reason may be lack of feature points in the low-textured scenes, so that many feature points last
only a few video frames which are easily neglected in key frame selection. The filter-based approach instead takes every
feature track into EKF update. Another reason is that we adopt novel information accumulation to get better triangulation
for long tracks as long as historical measurements outside the sliding window are discarded. The effectiveness of information accumulation of features is demonstrated in the next section.
\begin{figure}[h]
 \centering
 \includegraphics[width=0.36\textwidth]{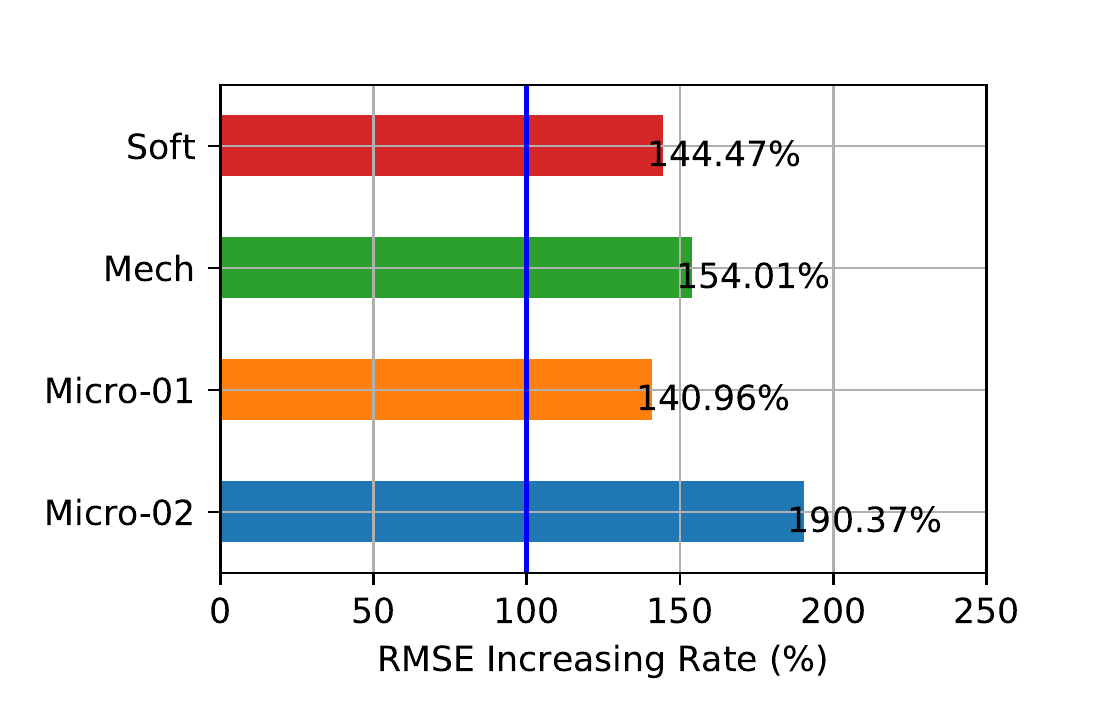}
 \caption{After disabling the information accumulation (both for points and lines), the RMSEs of StructVIO increase apparently about $157.45\%$ on average.}
 \label{fig:no_marg}
 \end{figure}
\rrev{\subsection{Information accumulation}}
\label{sec:exp_marg}
We evaluate the information accumulation method proposed in Section \ref{sec:marg}. In our implementation, we apply information accumulation to both point and line features. We check the performance difference of our StructVIO system using or without using information accumulation. As shown in Figure \ref{fig:no_marg}, if we disable information accumulation, the drift error of StructVIO increase about $150\%$. It suggests that information accumulation be helpful for long feature tracks to keep the historical information derived from the dropped measurements in old video frames. 

\subsection{Atlanta World v.s. Manhattan World}
We also conduct experiments to evaluate the advantage of using Atlanta world assumption instead of Manhattan world assumption. 
The tests were conducted on the 'Soft' and 'Mech' sequences, since both buildings contain oblique corridors or curvy structures  as shown in Figure \ref{fig:floor_plans}. 
As we can see, both scenes consists of two Manhattan worlds that are better described as a Atlanta world. To test the performance of using Manhattan world assumption,
we keep all the parameters the same disable detection of multiple Manhattan worlds. 
The benefit of using Atlanta world assumption instead of Manhattan world assumption as did in \cite{zhou2015structslam}  in such scenes is clearly shown in Table \ref{tab:maha_vs_atlan}. If we use Manhattan world assumption  instead of Atlanta world assumption in the two scenes with irregular structures, the RMSE errors increase about 
$287\% $ in average.

\rrev{\section{Discussion}
In this work, we mainly focus on feature-based visual-inertial odometry methods. The methods tested in the experiments are all feature-based VIO methods. We note that some visual inertial methods, like visual-inertial ORB-SLAM \cite{mur2017visual} and visual-inertial DSO \cite{von2018direct} report better results (in RMSE) than all tested methods including the proposed StructVIO method in the EuRoC datasets. However, visual-inertial ORB-SLAM is in fact a full SLAM system with map reuse and loop closing. The latter is a direct approach that operates at the pixel level without extracting features. 

Recent research \cite{forster2017svo}\cite{engel2018direct}\cite{von2018direct} shows that direct approach leads to higher accuracy than feature-based one in both benchmark tests and real world tests, but it is hard to say the direct approach is a solution better than feature-based approach since both of them have their advantages for particular applications. 
It is however really interesting to see that if the idea presented in this paper, validated by extensive tests, could be adopted in a direct approach to make a superior VIO system.

Another issue is about the failure cases of the proposed method. One may concern about the performance of our method when the scene is not as regular as the Atlanta world model describes. For example, the scene contains only slant walls or grounds, producing only slant line features.
Since our system adopts both points and line features, the slant lines are simply treated as outliers and our system acts as a point-only VIO. Our system may fail in the extreme case that the scene contains only slant walls without sufficient textures. Fortunately such kind of scenes may not usually happen.  

Our system is implemented in C++ without any optimization for computational time. We use the KLT features and tracker to detect and track the point features, and use LSD detector to detect line segments on the image. The whole pipeline is running in a single thread. The system runs on an i7 laptop about $15$ frames per second. The bottle neck is about line extraction and tracking (in fact our point-only system runs about $40$ frame per second). The performance of our MSCKF-based method however can be significantly improved. It has been shown that fast computation has be achieved for resourced-limited platforms in point-only MSCKF methods \cite{li2012vision}\cite{li2013real}. For example,  feature extraction (both points and lines) and tracking can be significantly sped up by using efficient algorithms (e.g. replacing KLT features with fast corners, detecting lines with Canny edges, and tracking with IMU predictions) and parallel processing (multi-thread or GPU processing). }

\section{Conclusion}
In this paper, we propose a novel visual-inertial navigation approach that integrates structural regularity of man-made environments by using line features with prior orientation. The structural regularity is modeled as an Atlanta world, which consists of multiple Manhattan worlds. The prior orientation is encoded in each local Manhattan world that is detected on-the-fly and is updated in the state variable over time. To realize such a visual-inertial navigation system , we made several technical contributions, including a novel parameterization for lines that integrate lines and Manhattan world together, a flexible strategy for detection and management of Manhattan worlds, a reliable line tracking method, and an information accumulation method for long line tracks. 

We compared our method with existing algorithms in both benchmark datasets and in real-world tests with a Project Tango tablet. The results show that our approach outperforms existing visual-inertial odometry methods that are considered as the state of the arts, though the test data are challenging because of lack of textures, bad lighting conditions and fast camera motions. That indicates incorporating structural regularity is helpful to implement a better visual-inertial system.


%




\ifCLASSOPTIONcaptionsoff
  \newpage
\fi


\bibliographystyle{IEEEtran}

\begin{thebibliography}{10}
\providecommand{\url}[1]{#1}
\csname url@rmstyle\endcsname
\providecommand{\newblock}{\relax}
\providecommand{\bibinfo}[2]{#2}
\providecommand\BIBentrySTDinterwordspacing{\spaceskip=0pt\relax}
\providecommand\BIBentryALTinterwordstretchfactor{4}
\providecommand\BIBentryALTinterwordspacing{\spaceskip=\fontdimen2\font plus
\BIBentryALTinterwordstretchfactor\fontdimen3\font minus
  \fontdimen4\font\relax}
\providecommand\BIBforeignlanguage[2]{{%
\expandafter\ifx\csname l@#1\endcsname\relax
\typeout{** WARNING: IEEEtran.bst: No hyphenation pattern has been}%
\typeout{** loaded for the language `#1'. Using the pattern for}%
\typeout{** the default language instead.}%
\else
\language=\csname l@#1\endcsname
\fi
#2}}

\bibitem{klein2007parallel}
G.~Klein and D.~Murray, ``Parallel tracking and mapping for small ar
  workspaces,'' in \emph{IEEE \& ACM Proc. of Int'l Sym. on Mixed and Augmented
  Reality}.\hskip 1em plus 0.5em minus 0.4em\relax IEEE, 2007, pp. 225--234.

\bibitem{mur2015orb}
R.~Mur-Artal, J.~Montiel, and J.~D. Tard{\'o}s, ``Orb-slam: a versatile and
  accurate monocular slam system,'' \emph{IEEE Trans. on Robotics}, vol.~31,
  no.~5, pp. 1147--1163, 2015.

\bibitem{zou2013coslam}
D.~Zou and P.~Tan, ``Coslam: Collaborative visual slam in dynamic
  environments,'' \emph{IEEE Trans. Pattern Analysis and Machine Intelligence},
  vol.~35, no.~2, pp. 354--366, 2013.

\bibitem{mourikis2007multi}
A.~I. Mourikis and S.~I. Roumeliotis, ``A multi-state constraint kalman filter
  for vision-aided inertial navigation,'' in \emph{Proc. IEEE Int. Conf.
  Robotics and Automation}.\hskip 1em plus 0.5em minus 0.4em\relax IEEE, 2007,
  pp. 3565--3572.

\bibitem{leutenegger2015keyframe}
S.~Leutenegger, S.~Lynen, M.~Bosse, R.~Siegwart, and P.~Furgale,
  ``Keyframe-based visual--inertial odometry using nonlinear optimization,''
  \emph{Int. J. Robotics Research}, vol.~34, no.~3, pp. 314--334, 2015.

\bibitem{qin2017vins}
T.~Qin, P.~Li, and S.~Shen, ``Vins-mono: A robust and versatile monocular
  visual-inertial state estimator,'' \emph{arXiv preprint arXiv:1708.03852},
  2017.

\bibitem{coughlan1999manhattan}
J.~M. Coughlan and A.~L. Yuille, ``Manhattan world: Compass direction from a
  single image by bayesian inference,'' in \emph{Proc. IEEE Int. Conf. Computer
  Vision}, vol.~2.\hskip 1em plus 0.5em minus 0.4em\relax IEEE, 1999, pp.
  941--947.

\bibitem{furukawa2009manhattan}
Y.~Furukawa, B.~Curless, S.~M. Seitz, and R.~Szeliski, ``Manhattan-world
  stereo,'' in \emph{Proc. IEEE Conf. Computer Vision \& Pattern
  Recognition}.\hskip 1em plus 0.5em minus 0.4em\relax IEEE, 2009, pp.
  1422--1429.

\bibitem{gupta2010blocks}
A.~Gupta, A.~A. Efros, and M.~Hebert, ``Blocks world revisited: Image
  understanding using qualitative geometry and mechanics,'' in \emph{Euro.
  Conf. Computer Vision}.\hskip 1em plus 0.5em minus 0.4em\relax Springer,
  2010, pp. 482--496.

\bibitem{ruotsalainen2012mitigation}
L.~Ruotsalainen, J.~Bancroft, and G.~Lachapelle, ``Mitigation of attitude and
  gyro errors through vision aiding,'' in \emph{IEEE Proc. of Indoor
  Positioning and Indoor Navigation}.\hskip 1em plus 0.5em minus 0.4em\relax
  IEEE, 2012, pp. 1--9.

\bibitem{zhou2015structslam}
H.~Zhou, D.~Zou, L.~Pei, R.~Ying, P.~Liu, and W.~Yu, ``Structslam: Visual slam
  with building structure lines,'' \emph{IEEE Trans. on Vehicular Tech.},
  vol.~64, no.~4, pp. 1364--1375, 2015.

\bibitem{schindler2004atlanta}
G.~Schindler and F.~Dellaert, ``Atlanta world: An expectation maximization
  framework for simultaneous low-level edge grouping and camera calibration in
  complex man-made environments,'' in \emph{Proc. IEEE Conf. Computer Vision \&
  Pattern Recognition}, vol.~1.\hskip 1em plus 0.5em minus 0.4em\relax IEEE,
  2004, pp. I--I.

\bibitem{smith2006real}
P.~Smith, I.~D. Reid, and A.~J. Davison, ``Real-time monocular slam with
  straight lines,'' in \emph{Proc. British Machine Vision Conf.}, 2006, pp.
  17--26.

\bibitem{sola2009undelayed}
J.~Sola, T.~Vidal-Calleja, and M.~Devy, ``Undelayed initialization of line
  segments in monocular slam,'' in \emph{IEEE/RSJ Proc. of Intelligent Robots
  and Systems}.\hskip 1em plus 0.5em minus 0.4em\relax IEEE, 2009, pp.
  1553--1558.

\bibitem{perdices2014lineslam}
E.~Perdices, L.~M. L{\'o}pez, and J.~M. Ca{\~n}as, ``Lineslam: Visual real time
  localization using lines and ukf,'' in \emph{ROBOT2013: First Iberian
  Robotics Conference}.\hskip 1em plus 0.5em minus 0.4em\relax Springer, 2014,
  pp. 663--678.

\bibitem{pumarola2017pl}
a.~pumarola, a.~vakhitov, a.~agudo, a.~sanfeliu, and f.~moreno noguer,
  ``pl-slam: real-time monocular visual slam with points and lines,'' in
  \emph{robotics and automation (icra), 2017 ieee international conference
  on}.\hskip 1em plus 0.5em minus 0.4em\relax ieee, 2017, pp. 4503--4508.

\bibitem{he2018pl}
Y.~He, J.~Zhao, Y.~Guo, W.~He, and K.~Yuan, ``Pl-vio: Tightly-coupled monocular
  visual--inertial odometry using point and line features,'' \emph{Sensors},
  vol.~18, no.~4, p. 1159, 2018.

\bibitem{sola2012impact}
J.~Sola, T.~Vidal-Calleja, J.~Civera, and J.~M.~M. Montiel, ``Impact of
  landmark parametrization on monocular ekf-slam with points and lines,''
  \emph{Int. J. Computer Vision}, vol.~97, no.~3, pp. 339--368, 2012.

\bibitem{gomez2016robust}
R.~Gomez-Ojeda and J.~Gonzalez-Jimenez, ``Robust stereo visual odometry through
  a probabilistic combination of points and line segments,'' in \emph{Robotics
  and Automation (ICRA), 2016 IEEE International Conference on}.\hskip 1em plus
  0.5em minus 0.4em\relax IEEE, 2016, pp. 2521--2526.

\bibitem{lee2009vpass}
Y.~H. Lee, C.~Nam, K.~Y. Lee, Y.~S. Li, S.~Y. Yeon, and N.~L. Doh, ``Vpass:
  Algorithmic compass using vanishing points in indoor environments,'' in
  \emph{IEEE/RSJ Proc. of Intelligent Robots and Systems}.\hskip 1em plus 0.5em
  minus 0.4em\relax IEEE, 2009, pp. 936--941.

\bibitem{zhang2012loop}
G.~Zhang, D.~H. Kang, and I.~H. Suh, ``Loop closure through vanishing points in
  a line-based monocular slam,'' in \emph{Proc. IEEE Int. Conf. Robotics and
  Automation}.\hskip 1em plus 0.5em minus 0.4em\relax IEEE, 2012, pp.
  4565--4570.

\bibitem{camposeco2015using}
F.~Camposeco and M.~Pollefeys, ``Using vanishing points to improve
  visual-inertial odometry,'' in \emph{Robotics and Automation (ICRA), 2015
  IEEE International Conference on}.\hskip 1em plus 0.5em minus 0.4em\relax
  IEEE, 2015, pp. 5219--5225.

\bibitem{kottas2013exploiting}
D.~G. Kottas and S.~I. Roumeliotis, ``Exploiting urban scenes for vision-aided
  inertial navigation.'' in \emph{Robotics: Science and Systems}, 2013.

\bibitem{civera2008inverse}
J.~Civera, A.~J. Davison, and J.~M. Montiel, ``Inverse depth parametrization
  for monocular slam,'' \emph{IEEE Trans. on Robotics}, vol.~24, no.~5, pp.
  932--945, 2008.

\bibitem{sola2017quaternion}
J.~Sola, ``Quaternion kinematics for the error-state kalman filter,''
  \emph{arXiv preprint arXiv:1711.02508}, 2017.

\bibitem{von2010lsd}
R.~G. Von~Gioi, J.~Jakubowicz, J.-M. Morel, and G.~Randall, ``Lsd: A fast line
  segment detector with a false detection control,'' \emph{IEEE Trans. Pattern
  Analysis and Machine Intelligence}, vol.~32, no.~4, pp. 722--732, 2010.

\bibitem{li2013optimization}
M.~Li and A.~I. Mourikis, ``Optimization-based estimator design for
  vision-aided inertial navigation,'' in \emph{Robotics: Science and Systems},
  2013, pp. 241--248.

\bibitem{davison2007monoslam}
A.~J. Davison, I.~D. Reid, N.~D. Molton, and O.~Stasse, ``Monoslam: Real-time
  single camera slam,'' \emph{IEEE Trans. Pattern Analysis and Machine
  Intelligence}, vol.~29, no.~6, pp. 1052--1067, 2007.

\bibitem{toldo2008robust}
R.~Toldo and A.~Fusiello, ``Robust multiple structures estimation with
  j-linkage,'' in \emph{Euro. Conf. Computer Vision}.\hskip 1em plus 0.5em
  minus 0.4em\relax Springer, 2008, pp. 537--547.

\bibitem{burri2016euroc}
M.~Burri, J.~Nikolic, P.~Gohl, T.~Schneider, J.~Rehder, S.~Omari, M.~W.
  Achtelik, and R.~Siegwart, ``The euroc micro aerial vehicle datasets,''
  \emph{The International Journal of Robotics Research}, vol.~35, no.~10, pp.
  1157--1163, 2016.

\bibitem{furgale2013unified}
P.~Furgale, J.~Rehder, and R.~Siegwart, ``Unified temporal and spatial
  calibration for multi-sensor systems,'' in \emph{IEEE/RSJ Proc. of
  Intelligent Robots and Systems}.\hskip 1em plus 0.5em minus 0.4em\relax IEEE,
  2013, pp. 1280--1286.

\bibitem{de2015aruco}
A.~de~la Visi{\'o}n~Artificial, ``Aruco. a minimal library for augmented
  reality applications based on opencv,'' \emph{Dosegljivo: http://www. uco.
  es/investiga/grupos/ava/node/26.[Dostopano: 16. 4. 2016]}, 2015.

\bibitem{devernay2001straight}
F.~Devernay and O.~Faugeras, ``Straight lines have to be straight,''
  \emph{Machine vision and applications}, vol.~13, no.~1, pp. 14--24, 2001.

\end{thebibliography}

\end{document}